\documentclass[letterpaper]{article} 
\usepackage{aaai2027}  
\usepackage[hyphens]{url}  
\usepackage{graphicx} 
\usepackage{natbib}  
\usepackage{caption} 
\usepackage{subcaption} 
\usepackage{amssymb,amsmath}
\usepackage{mathtools}

\usepackage{listings} 
\usepackage{algorithm}
\usepackage{algorithmic}
\usepackage{multirow}
\usepackage{array}
\usepackage{makecell}
\usepackage{booktabs}
\usepackage[table]{xcolor}
\usepackage{arydshln}
\usepackage{xcolor}
\usepackage{amsthm}
\usepackage{tikz}
\usepackage{fvextra}
\usetikzlibrary{positioning,arrows.meta}

\newcommand{\up}[1]{{\color{red}#1$\uparrow$}}
\newcommand{\down}[1]{{\color{green!60!black}#1$\downarrow$}}
\definecolor{delogblue}{HTML}{EAF3FF}
\definecolor{proprietarygray}{gray}{0.93}

\usepackage{newfloat}
\usepackage{listings}
\DeclareCaptionStyle{ruled}{labelfont=normalfont,labelsep=colon,strut=off} 
\floatstyle{ruled}
\newfloat{listing}{tb}{lst}{}
\floatname{listing}{Listing}

\usepackage{xcolor}
\usepackage[most]{tcolorbox}
\tcbuselibrary{listings,breakable}

\newtcblisting{GraphPromptBox}{
    enhanced,
    breakable,
    listing only,
    colback=gray!3,
    colframe=black!65,
    boxrule=0.5pt,
    arc=1mm,
    left=2mm,
    right=2mm,
    top=1mm,
    bottom=1mm,
    title={Prompt for Cognitive Attribution Map Construction},
    colbacktitle=black!65,
    coltitle=white,
    fonttitle=\bfseries\small,
    listing options={
      basicstyle=\ttfamily\scriptsize,
      breaklines=true,
      numbers=none,
      breakatwhitespace=false,
      columns=fullflexible,
      keepspaces=true,
      showstringspaces=false,
      tabsize=2,
      literate=
        {é}{{\'e}}1
        {–}{{--}}1
        {（}{{(}}1
        {）}{{)}}1
    }
  }

\newtcblisting{AnswerPromptBox}{
    enhanced,
    breakable,
    listing only,
    colback=gray!3,
    colframe=black!65,
    boxrule=0.5pt,
    arc=1mm,
    left=2mm,
    right=2mm,
    top=1mm,
    bottom=1mm,
    title={Prompt for Long-form Answer Generation on the Map},
    colbacktitle=black!65,
    coltitle=white,
    fonttitle=\bfseries\small,
    listing options={
      basicstyle=\ttfamily\scriptsize,
      breaklines=true,
      numbers=none,
      breakatwhitespace=false,
      columns=fullflexible,
      keepspaces=true,
      showstringspaces=false,
      tabsize=2,
      literate=
        {é}{{\'e}}1
        {–}{{--}}1
        {（}{{(}}1
        {）}{{)}}1
    }
  }

\lstdefinelanguage{Datalog}{
  alsoletter={_},
  morekeywords={
    id,name, content, Nodes, Edges,subgraph0, nodes, inputs, outputs, subgraph_description
  },
  morestring=[b]",
  sensitive=true
}
\definecolor{logicblue}{RGB}{54, 112, 178}
\definecolor{logicback}{RGB}{247, 250, 253}
\definecolor{logicframe}{RGB}{166, 190, 216}
\definecolor{logictitle}{RGB}{232, 241, 250}

\usepackage[most]{tcolorbox}
\tcbuselibrary{listings,breakable,skins}

\definecolor{logicback}{RGB}{248,249,250}
\definecolor{logicframe}{RGB}{180,180,180}
\definecolor{logictitle}{RGB}{238,241,245}
\definecolor{logicblue}{RGB}{35,85,140}

\newtcblisting[auto counter]{logicprogram}[2]{
  enhanced,
  breakable,
  listing only,
  colback=logicback,
  colframe=logicframe,
  colbacktitle=logictitle,
  coltitle=black,
  boxrule=0.3pt,
  arc=1pt,
  left=3pt,
  right=3pt,
  top=2pt,
  bottom=2pt,
  boxsep=1pt,
  title={Example graph~\thetcbcounter: #2},
  fonttitle=\bfseries\scriptsize,
  titlerule=0.25pt,
  toptitle=1pt,
  bottomtitle=1pt,
  lefttitle=3pt,
  righttitle=3pt,
  label={#1},
  listing options={
    language=Datalog,
    basicstyle=\fontfamily{pcr}\fontsize{6.8pt}{7.2pt}\selectfont,
    keywordstyle=\color{logicblue!75!black}\bfseries,
    stringstyle=\color{black!75},
    commentstyle=\color{black!55},
    columns=flexible,
    keepspaces=false,
    breaklines=true,
    breakatwhitespace=false,
    breakautoindent=false,
    breakindent=0pt,
    postbreak={},
    showstringspaces=false,
    numbers=none,
    frame=none,
    xleftmargin=0pt,
    aboveskip=0pt,
    belowskip=0pt,
    lineskip=-1pt,
    tabsize=2
  },
  before skip=4pt,
  after skip=4pt
}

\title{CAGE: Cognitive Attribution Graphs for Faithful Inline Citation Generation in Long-Form Question Answering}
\author{
Zhichao Yan\textsuperscript{\rm 1},
Shizhao Li\textsuperscript{\rm 1},
Jiapu Wang\textsuperscript{\rm 2},
Haoran Luo\textsuperscript{\rm 4},
Qingang Zhang\textsuperscript{\rm 6},
Jiaoyan Chen\textsuperscript{\rm 3},\\
Ru Li\textsuperscript{\rm 1},
Jeff Z. Pan\textsuperscript{\rm 5}
}

\affiliations{
\textsuperscript{\rm 1}Shanxi University, Taiyuan, China\\
\textsuperscript{\rm 2}Nanjing University of Science and Technology, Nanjing, China\\
\textsuperscript{\rm 3}The University of Manchester, Manchester, UK\\
\textsuperscript{\rm 4}Nanyang Technological University, Singapore\\
\textsuperscript{\rm 5}The University of Edinburgh, Edinburgh, UK\\
\textsuperscript{\rm 6}Jilin University, Jilin, China\\
\texttt{\{yanzhichao,lishizhao,liru\}@sxu.edu.cn}\\
\texttt{jiapu.wang@njust.edu.cn},
\texttt{jiaoyan.chen@manchester.ac.uk}\\
\texttt{qinggangzhang@jlu.edu.cn}, \texttt{j.z.pan@ed.ac.uk}
}

\begin{document}

\maketitle

\begin{abstract}
Long-form question answering increasingly relies on retrieved evidence to make
LLM outputs verifiable, with inline citations tracing claims to source
documents. However, existing systems often attach citations that are
topically related but insufficient to support their claims. We identify
\emph{attribution ambiguity} as a structural challenge: end-to-end generation
must implicitly resolve combinatorial claim--document assignments, obscuring
evidential boundaries and increasing the risk of
\emph{evidence-boundary overrun}, where claims exceed cited support. To address
this challenge, we propose \textsc{CAGE} (\textbf{\underline{C}}ognitive
\textbf{\underline{A}}ttribution \textbf{\underline{G}}raphs for Citation
G\textbf{\underline{e}}neration), a two-stage framework that introduces an
explicit cognitive attribution map before answer generation. CAGE first trains
a plug-and-play Cognitive Map Induction Model to construct answer-centered
support subgraphs, aligning each semantic answer unit with supporting documents
through explicit relations. A Structured Citation Reasoning Model then realizes
these units as sentence-level claims with map-aligned citations. Experiments on
ASQA, ELI5, and ExpertQA show that CAGE achieves state-of-the-art performance,
demonstrating the effectiveness of attribution-space contraction and
map-guided citation generation.
\end{abstract}


\section{Introduction}

Large language models (LLMs) can generate fluent long-form responses for knowledge-intensive tasks, yet hallucinations and unreliable source attribution undermine the verifiability and trustworthiness of their outputs \cite{farquhar2024detecting,yan2025atomic,chen2024complex,wang2023survey}. Retrieval-augmented generation (RAG) partially mitigates this issue by grounding generation in retrieved documents \cite{10387715,pan2023large,cheng2026graphrag,zhuo2025effective,xu2025log}, but document grounding alone does not establish whether each individual claim is supported by sufficient evidence. This limitation is especially consequential in downstream applications that consume LLM outputs without independent verification \cite{zhang2026chaining,luo2025large,wang2024large}, making explicit claim-level attribution essential for trustworthy long-form generation.

Citation-grounded long-form generation requires claims to include citations to
supporting documents \cite{Gao2022RARRRA,bohnet2022attributed}. Yet when claims
and citations are generated jointly, the model must simultaneously decide what
to state and what evidence to cite without explicit claim-specific support
relations. Under weak, incomplete, or distributed evidence, this coupled process
may produce claims that exceed available support while attaching topically
related but insufficient citations
\cite{sun2025redeep,qian2026relevant,10908675}. We call this
\emph{evidence-boundary overrun}, where a claim exceeds what its cited documents
can strictly justify.

\begin{figure}[t]
\centering
\includegraphics[width=1\linewidth]{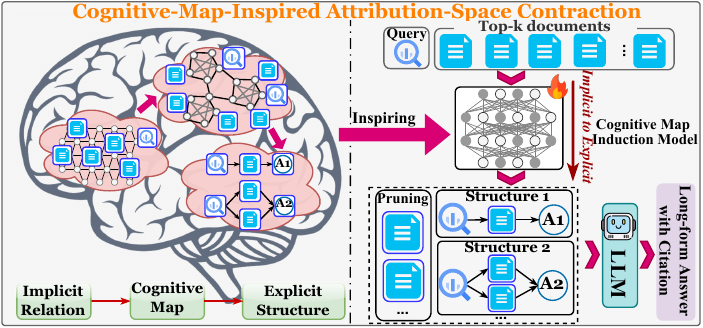}
\caption{
Motivation of CAGE. The human brain abstracts scattered observations into relational cognitive maps to organize and associate discrete information. CAGE simulates this workflow with a plug-and-play Cognitive Map Induction Model that converts retrieved documents into intermediate evidence structures, contracting the attribution space before LLM-based citation generation.
}
\label{fig:introduction}
\end{figure}

The risk of such overrun is amplified by \emph{attribution ambiguity}:
retrieved documents may be relevant to the question without specifying which
subset supports each individual claim. A document may provide only partial
evidence, while different claims may require different document combinations.
For a fixed set of \(m\) candidate claims and \(n\) retrieved documents, each
claim may be assigned any non-empty document subset, yielding up to
\((2^n-1)^m\) possible citation assignments. This combinatorial space does not
itself imply unsupported generation, but resolving it implicitly during
decoding makes it difficult to distinguish topical relevance from sufficient
support. Explicitly determining claim-specific support relations before final generation can therefore improve the verifiability of citation-grounded outputs and reduce the risk of \emph{evidence-boundary overrun}.

Cognitive-map theories suggest that complex reasoning can benefit from
transforming implicit relations among observations into explicit relational
structures for structured inference
\cite{mark2020transferring,whittington2022build,tan2025medial}. Drawing on this
perspective, citation-grounded generation should externalize latent relations
between retrieved evidence and semantic answer units as an explicit attribution
structure before final decoding. Such a structure can exclude invalid evidence
assignments, organize supporting documents around answer units, and make
claim-specific support relations explicit. This also explains why stronger
prompting or larger generators may still produce missing or incorrect
citations: \textit{improving model capacity alone may be insufficient without an explicit
mechanism that externalizes claim--document relations and constrains
unsupported assignments before generation.}

To this end, we introduce \textsc{CAGE}
(\textbf{\underline{C}}ognitive \textbf{\underline{A}}ttribution
\textbf{\underline{G}}raphs for Citation G\textbf{\underline{e}}neration), a
two-stage framework for inline citation generation. As illustrated in Figure~\ref{fig:introduction}, CAGE first trains a plug-and-play
\emph{Cognitive Attribution Map Induction Model} to construct a cognitive attribution map
from a question and its retrieved documents. The map consists of
answer-centered support subgraphs that link answer units to their supporting documents and explicitly define the evidence boundary; unsupported documents are excluded, while the absence of supported answer units triggers evidence-driven abstention. In the second stage, a \emph{Structured Citation Reasoning Model} reasons over
the induced map, realizes its semantic answer nodes as coherent sentence-level
claims, and generates inline citations aligned with the corresponding graph-specified document identifiers. CAGE externalizes answer--document attribution as validated support
subgraphs, decoupling attribution from generation across heterogeneous
backbones.

In summary, the main contributions are as follows:
\begin{itemize}
\item We identify \emph{attribution ambiguity} as a structural cause of
citation error in end-to-end generation, where unconstrained
claim--document assignment induces a combinatorial citation-assignment space
and allows topical relevance to substitute for evidential support.

\item We propose \textsc{CAGE}, a two-stage framework that trains a
plug-and-play Cognitive Attribution Map Induction Model and a Structured Citation Reasoning
Model to explicitly determine semantic answer units and their supporting
documents before final answer realization.

\item We demonstrate best performance on ASQA, ELI5, and
ExpertQA, and show that CAGE consistently improves proprietary LLMs,
indicating that cognitive attribution map provide complementary gains beyond
document filtering and enhance strong end-to-end systems.
\end{itemize}

\section{Preliminaries}
\label{preliminary}

\subsection{Task Definition: Inline Citation Generation}
\label{sec:task-definition}

Given a question \(Q\) and top-\(k\) retrieved documents
\(\mathcal{D}_k=\{d_i\}_{i=1}^{k}\), inline citation generation requires an
LLM \(\mathcal{M}\) to produce a long-form answer grounded in specific source
documents. We formalize the output as
\begin{equation}
    \mathcal{Y}
    =
    \{(s_j,c_j)\}_{j=1}^{m}
    \leftarrow
    \mathcal{M}(Q,\mathcal{D}_k),
\end{equation}
where $s_j$ is an individual factual statement and $c_j\subseteq\mathcal{D}_k$ is the citation set supporting it, \(m\) is the number of claims. If $\mathcal{D}_k$ lacks sufficient evidence for $Q$, the model should perform evidence-driven abstention by returning an explicit refusal.

\subsection{Cognitive Attribution Map}
\label{sec:prelim-support-graph}
Given a query \(Q\) and \(\mathcal{D}_k=\{d_1,\ldots,d_k\}\), the map \(\mathbf{G}\) is defined as a collection of answer-centered support subgraphs:
\begin{equation}
    \mathbf{G}=\{\mathcal{G}_r\}_{r=1}^{m},
    \qquad
    0\leq m.
\end{equation}
Each subgraph represents one semantic answer unit:
\begin{equation}
\mathcal{G}_r
=
(\mathcal{V}_r,\mathcal{E}_r;\rho_r),
\qquad
\mathcal{V}_r
=
\left\{Q\right\}\cup\mathcal{D}_r\cup \left\{a_r\right\},
\label{node_schema}
\end{equation}
where \(\mathcal{D}_r\subseteq\mathcal{D}_k\) is a non-empty supporting
document set, \(a_r\) is a canonical semantic answer unit supported by
\(\mathcal{D}_r\), and \(\rho_r\) describes their support relation. The answer
node specifies the content to be conveyed without fixing its final wording or
position in the long-form answer.

A document node
\(d_i=(\mathrm{id},\mathrm{name},\mathrm{content})\) stores its citation ID,
title, and evidence text. The query node follows the same schema, with
\(\mathrm{id}(Q)=\texttt{Q}\), \(\mathrm{name}(Q)\) containing the query, and \(\mathrm{content}(Q)\) a fixed instruction. An answer node
\begin{equation}
a_r
=
\bigl(
\mathrm{id}(a_r), \mathrm{name}(a_r),
\mathrm{content}(a_r)
\bigr),
\end{equation}
uses a unique unit ID, the fixed name ``Answer Unit'', and its canonical semantics as content. The edge set is
\begin{equation}
\mathcal{E}_r
=
\{
(Q,d_i),(d_i,a_r)
\mid d_i\in\mathcal{D}_r
\},
\end{equation}
encodes evidence selection through query-to-document edges and evidential support through document-to-answer edges. A complete subgraph example is provided below:
\begin{logicprogram}{prog:interleaved}{}
 Nodes: id:Q   name: Who wrote The Sound of Silence?   content: Reasoning the graph with node A based on the node starting with d.
  id:d1   name: The Sound of Silence   content: "The Sound of Silence" is a song by Simon & Garfunkel. It was written by Paul Simon.
  id:A1 name: Answer Unit content: Paul Simon wrote "The Sound of Silence".
Edges: nodes:Q  inputs:      outputs: d1
  nodes:d1 inputs: Q    outputs: A1
  nodes:A1 inputs: d1   outputs: N/A
subgraph_description: The document supports the answer by stating that "The Sound of Silence" was written by Paul Simon.
\end{logicprogram}

\paragraph{Attribution-space contraction.}
The graph induces the retained document set
\(\mathcal{D}^{+}=\bigcup_{r=1}^{m}\mathcal{D}r\), with
\(n=|\mathcal{D}^{+}|\), and specifies the attribution target
\begin{equation}
\alpha_{\mathbf{G}}(a_r)
=
\left\{
\mathrm{id}(d_i)
\mid
d_i\in\mathcal{D}_r
\right\}.
\end{equation}
Thus, if a generated claim \(s{\pi(r)}\) semantically realizes \(a_r\), its
target citation set is
\(c_{\pi(r)}^{*}=\alpha_{\mathbf{G}}(a_r)\). Documents outside
\(\mathcal{D}^{+}\) are excluded from generation; if \(m=0\), CAGE abstains.

\begin{figure*}[t]
    \centering
    \includegraphics[width=\textwidth]{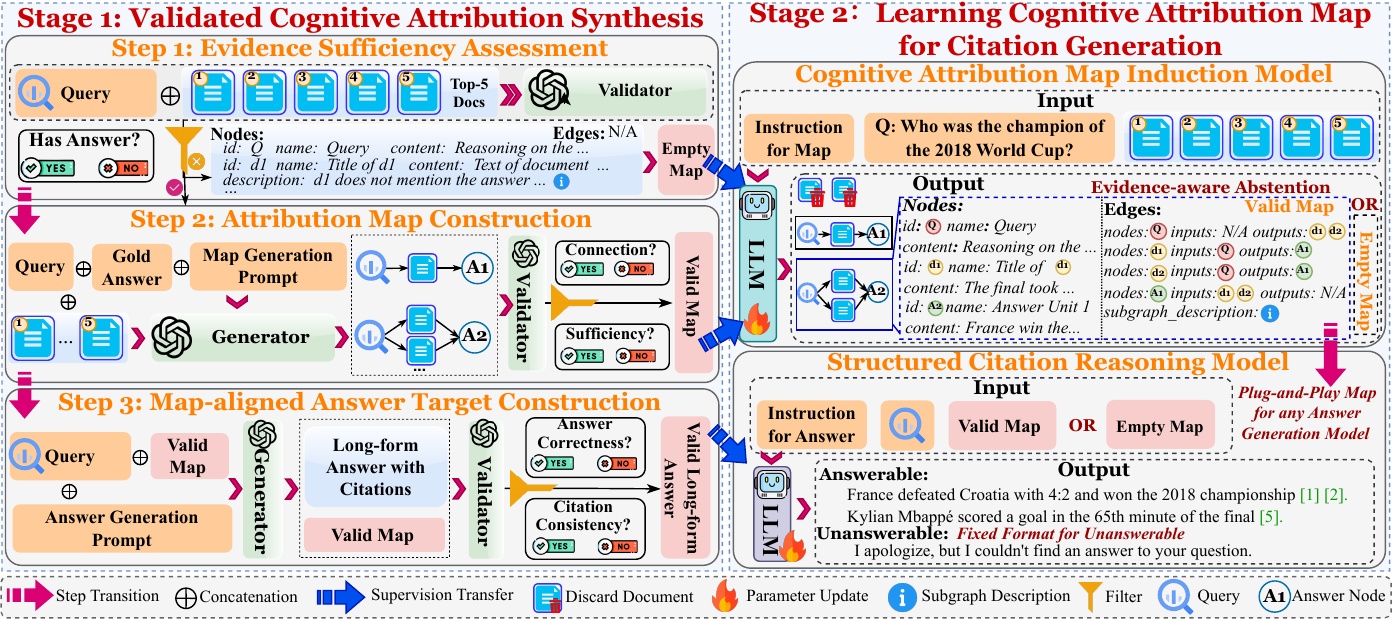}
    \caption{
    Overview of \textsc{CAGE}. Stage 1 Validated Cognitive Attribution Synthesis constructs
    validated cognitive attribution map and map-aligned answer supervision,
    which are used to train the Cognitive Attribution Map Induction Model and the Structured
    Citation Reasoning Model, respectively, in Stage 2.
    }
    \label{fig:main}
\end{figure*}

Without an attribution map, each of the \(m\) claims may select any non-empty
subset of the \(n\) retained documents, yielding up to
\begin{equation}
|\Omega_{\mathrm{unstr}}|
=
(2^n-1)^m .
\end{equation}
Once induced, the attribution map \(\mathbf{G}\) fixes one citation set for
each answer node:
\begin{equation}
\Omega_{\mathbf{G}}
=
\left\{
\big(
\alpha_{\mathbf{G}}(a_1),\ldots,
\alpha_{\mathbf{G}}(a_m)
\big)
\right\},
\qquad
|\Omega_{\mathbf{G}}|=1 .
\end{equation}
Thus, conditioned on the induced map, CAGE reduces the downstream
citation-assignment space from up to \((2^n-1)^m\) unconstrained choices to
one graph-specified target, while attribution is resolved during map
induction.

\section{Methodology}
\label{sec:method}

This section presents \textsc{CAGE}, a two-stage framework for
citation-grounded long-form answer generation. \textit{Validated Cognitive Attribution Synthesis}
(VCAS) constructs validated cognitive attribution map and map-aligned answers,
which supervise a \textit{Cognitive Attribution Map Induction Model} (CMI) for constructing
answer-centered support subgraphs and a Structured Citation Reasoning Model
(SCR) for map-conditioned answer generation.

\subsection{Validated Cognitive Attribution Synthesis (VCAS)}
\label{sec:VCAS}

This section introduces how VCAS constructs supervision through two LLM-instantiated roles: a generator and a validator. As shown in Figure~\ref{fig:main}, VCAS converts query--document instances into two aligned supervision sets: one for CMI and the other for SCR.

Since LLMs generate token sequences, a non-empty cognitive attribution map
\(\mathbf{G}=\{\mathcal{G}r\}_{r=1}^{m}\) is represented by a canonical
textual encoding:
\begin{equation}
\begin{aligned}
z_{\mathrm{map}}
&=
\operatorname{Enc}_{\mathrm{map}}(\mathbf{G}) \\
&=
\operatorname{Enc}_{V}(\mathcal{V}(\mathbf{G}))
\oplus
\operatorname{Enc}_{E}(\mathcal{E}(\mathbf{G}))
\oplus
\operatorname{Enc}_{\rho}(\boldsymbol{\rho}(\mathbf{G})),
\end{aligned}
\end{equation}
where \(\mathcal{V}(\mathbf{G})\), \(\mathcal{E}(\mathbf{G})\), and
\(\boldsymbol{\rho}(\mathbf{G})\) denote its nodes, edges, and
descriptions, respectively. The encoding preserves the map structure while
serving as an autoregressive prediction target.

\textit{Step 1: Evidence Sufficiency Assessment.}
Given a query \(Q\) and its top-\(k\) retrieved documents
\(\mathcal{D}_k=\{d_1,\ldots,d_k\}\), the validator determines whether the retrieved documents contain enough evidence to support at least one query-responsive answer unit:
\begin{equation}
\exists a,\ \exists\mathcal{D}'\subseteq\mathcal{D}_k,
\qquad
\mathcal{D}'\neq\emptyset,
\qquad
\mathcal{D}'\models_Q a .
\end{equation}
Here, \(\mathcal{D}'\models_Q a\) means that \(\mathcal{D}'\) provides
sufficient evidence for \(a\) under \(Q\).

Unsupported instances are assigned a document preserving empty map:
\begin{equation}
\begin{aligned}
\mathbf{G}^{*}_{\emptyset}
&=
(\mathcal{V}^{*}_{\emptyset},
\mathcal{E}^{*}_{\emptyset};
\rho^{*}_{\emptyset}),\\
\mathcal{V}^{*}_{\emptyset}
&=
d_i,
\quad
\mathcal{E}^{*}_{\emptyset}
=
\emptyset,
\quad
z^{*}_{\mathrm{map},\emptyset}
=
\operatorname{Enc}_{\mathrm{map}}
(\mathbf{G}^{*}_{\emptyset}).
\end{aligned}
\label{eq}
\end{equation}
Here, \(\rho^{*}_{\emptyset}\) explains the evidence insufficiency. The empty map retains all candidate document nodes but contains no answer nodes, selected supporting-document assignments, or support edges. These instances train CMI to identify
unsupported questions without constructing spurious answer units. The seed questions are sampled from the training sets of FRONT
\cite{huang2024learning} and ASQA \cite{stelmakh-etal-2022-asqa} and the number of documents was set to 5, consistent with prior study \cite{song2025measuring}.

\textit{Step 2: Attribution Map Construction.}
For each supported instance, the generator
\(\mathcal{M}_{\mathrm{gen}}\) receives \((Q,\mathcal{D}_k, \mathcal{P_\mathrm{map}})\) and generates a
candidate cognitive attribution map through its canonical textual encoding:
\begin{equation}
\begin{aligned}
    \overline{z}_{\mathrm{map}}
    &=
    \mathcal{M}_{\mathrm{gen}}(Q,\mathcal{D}_k, \mathcal{P_\mathrm{map}}) \\
    &=
    \operatorname{Enc}_{\mathrm{map}}
    \bigl(\overline{\mathbf{G}}\bigr),
    \qquad
    \overline{\mathbf{G}}
    =
    \left\{
        \overline{\mathcal{G}}_r
    \right\}_{r=1}^{\bar{m}} .
\end{aligned}
\end{equation}
where \(\mathcal{P_\mathrm{map}}\) is a specific prompt for map generation.

The validator checks each candidate subgraph for structural connectivity and
evidential support:
\begin{equation}
    \mathbb{V}_Q
    \bigl(\overline{\mathcal{G}}_r\bigr)
    =
    \mathbb{I}
    \left[
        \operatorname{Connected}
        \bigl(\overline{\mathcal{G}}_r\bigr)
        \land
        \overline{\mathcal{D}}_r
        \models_Q
        \overline{a}_r
    \right].
\end{equation}
Here, \(\operatorname{Connected}(\overline{\mathcal{G}}_r)\) requires
\(\overline{\mathcal{D}}_r\neq\emptyset\) and every document node to lie on a
complete support path from the query node to the semantic answer node:
\begin{equation}
    \forall d_i\in\overline{\mathcal{D}}_r,\qquad
    (Q,d_i)\in\overline{\mathcal{E}}_r
    \ \land\
    (d_i,\overline{a}_r)\in\overline{\mathcal{E}}_r.
\end{equation}
Thus, all query, document, and answer nodes participate in the support
structure, with no isolated nodes. The relation
\(\overline{\mathcal{D}}_r\models_Q\overline{a}_r\) indicates that the
assigned documents collectively provide sufficient evidence for the semantic
answer node.

The accepted subgraphs are assembled into the validated cognitive attribution
map:
\begin{equation}
\mathbf{G}^{*}
=
\{
\overline{\mathcal{G}}_r
\in
\overline{\mathbf{G}}
\mid
\mathbb{V}_Q
\bigl(\overline{\mathcal{G}}_r\bigr)
=1
\}.
\end{equation}
A supported instance is discarded if \(\mathbf{G}^{*}=\emptyset\).
Otherwise, the validated map is encoded as the CMI target:
\begin{equation}
z_{\mathrm{map}}^{*}
=
\operatorname{Enc}_{\mathrm{map}}
\bigl(\mathbf{G}^{*}\bigr).
\label{eq:VCAS-graph-target}
\end{equation}

Together with unsupported instances encoded as document preserving empty maps, these targets form the CMI supervision set
\(\mathcal{S}_{\mathrm{CMI}}\), which contains 7.8K samples.

\textit{Step 3: Map-aligned Answer Target Construction.} Since each retained document node contains its document ID, title, and textual
content, the encoded map provides both textual evidence and citation identifiers
for answer generation. The generator receives \((Q,z_\mathrm{map}^{*},\mathcal{P_\mathrm{ans}})\) and constructs a
map-aligned long-form answer with inline citations:
\begin{equation}
    \mathcal{Y}^{*}
    =
    \{(s^{*}_j,c^{*}_j)\}_{j=1}^{m}=\mathcal{M}_\mathrm{ans}(Q,z^*_\mathrm{map},\mathcal{P}_\mathrm{ans}),
\end{equation}
where \(s^{*}_j\) is a sentence-level claim and \(c^{*}_j\) is its citation set, \(\mathcal{P}_\mathrm{ans}\) is a prompt for map-aligned answer generation.

The answer target must satisfy a one-to-one alignment between graph nodes and generated claims. Since the final answer order need not follow the graph order, correctness is defined by the existence of a permutation \(\pi\) such that
\begin{equation}
    s^{*}_{\pi(r)}
    \equiv
    a^{*}_r,
    c^{*}_{\pi(r)}
    =
    \alpha_{\mathbf{G}^{*}}(a^{*}_r),
    \forall r\in\{1,\ldots,m\}.
\label{eq:target-alignment}
\end{equation}
Thus, every generated
claim must realize exactly one validated answer node, and its citation set must match the citation target specified by the corresponding answer-centered support subgraph.

The validator retains only answers that are correct with respect to the validated answer nodes and whose citation assignments are consistent with the graph-specified mapping. The accepted outputs form \(\mathcal{S}_{\mathrm{SCR}}\), containing 2.3K instances. The prompt of \(\mathcal{P_\mathrm{map}}\) and \(\mathcal{P_\mathrm{ans}}\) can be found in Appendix.

\subsection{Learning Cognitive Attribution Map for Citation Generation}
\label{sec:dual-stage-training}

CAGE factorizes citation-grounded generation through the textual encoding of a cognitive attribution map:
\begin{equation}
\begin{aligned}
&p_{\theta_g,\theta_a}\left(z_{\mathrm{map}},\mathcal{Y}\mid Q,\mathcal{D}_k,\mathcal{P_\mathrm{map}},\mathcal{P_\mathrm{ans}}\right)\\
&\qquad =p_{\theta_g}\left(z_{\mathrm{map}}\mid Q,\mathcal{D}_k,\mathcal{P_\mathrm{map}}\right)p_{\theta_a}\left(\mathcal{Y}\mid Q,z_{\mathrm{map}},\mathcal{P_\mathrm{ans}}\right).
\end{aligned}
\label{eq}
\end{equation}
Inference follows the same two-stage decomposition:
\begin{equation}
\begin{aligned}
\hat z_{\mathrm{map}}
&=\arg\max_{z}p_{\theta_g}(z\mid Q,\mathcal{D}_k,\mathcal{P_\mathrm{map}}),\\
\hat{\mathcal{Y}}
&=\arg\max_{\mathcal{Y}}p_{\theta_a}(\mathcal{Y}\mid Q,\hat z_{\mathrm{map}},\mathcal{P_\mathrm{ans}}).\end{aligned}\end{equation}

\begin{table*}[t]
\centering
\scriptsize
\setlength{\tabcolsep}{3pt} 
\renewcommand{\arraystretch}{0.85}
\begin{tabular}{
cc
cccc
cccc
cccc
c
}
\hline
\multirow{3}{*}{\textbf{SCR Model}} & \multirow{3}{*}{\textbf{Method}} 
& \multicolumn{4}{c}{\makecell{ASQA(610 answerable, 338 unanswerable)}}
& \multicolumn{4}{c}{\makecell{ExpertQA(682 answerable, 1487 unanswerable)}} 
& \multicolumn{4}{c}{\makecell{ELI5(207 answerable, 793 unanswerable)}}
& \multirow{3}{*}{\textbf{Average}}  \\

\cmidrule{3-14}

& & $\mathrm{EM}^\text{F1}_{\mathrm{AC}}$ & F1$_\text{GR}$ & F1$_\text{GC}$ & \textbf{TRUST}
  & $\mathrm{EM}^\text{F1}_{\mathrm{AC}}$ & F1$_\text{GR}$ & F1$_\text{GC}$ & \textbf{TRUST}
  & $\mathrm{EM}^\text{F1}_{\mathrm{AC}}$ & F1$_\text{GR}$ & F1$_\text{GC}$ & \textbf{TRUST}\\
  
\hline

\multirow{7}{*}{Qwen-2.5-0.5B}
& ICL\(^\dagger\) 
& 20.96 & 47.19 & 0.34 & \cellcolor{delogblue}22.83
& 21.42 & 38.71 & 0.44 & \cellcolor{delogblue}20.19
& 13.73 & 33.14 & 0.37 & \cellcolor{delogblue}15.75 & \cellcolor{proprietarygray}19.59 \\

& PostCite\(^\dagger\) 
& 8.55 & 50.84 & 8.23 & \cellcolor{delogblue}22.54
& 13.32 & 48.08 & 5.60 & \cellcolor{delogblue}22.33
& 9.87 & 27.10 & 4.10 & \cellcolor{delogblue}13.69 & \cellcolor{proprietarygray}19.52\\

& PostAttr 
& 8.55 & 50.84 & 2.23 & \cellcolor{delogblue}20.54
& 13.32 & 48.08 & 1.49 & \cellcolor{delogblue}20.96
& 9.87 & 27.10 & 0.68 & \cellcolor{delogblue}12.55 & \cellcolor{proprietarygray}18.02\\

& FRONT\(^\dagger\) 
& 42.83 & 39.15 & 45.87 & \cellcolor{delogblue}42.62
& 18.27 & 24.05 & 34.62 & \cellcolor{delogblue}25.65
& 13.74 & 17.29 & 27.95 & \cellcolor{delogblue}19.66 & \cellcolor{proprietarygray}29.31\\

& T-ALIGN\(^\dagger\) 
& 50.59 & 61.28 & 52.40 & \cellcolor{delogblue}54.76
& 18.16 & 63.31 & 35.07 & \cellcolor{delogblue}38.85
& 13.68 & 60.79 & 22.72 & \cellcolor{delogblue}32.40 & \cellcolor{proprietarygray}42.00\\
\cdashline{2-14}
& \textsc{CAGE} 
& \textbf{56.90} & \textbf{69.89} & \textbf{70.49} & \cellcolor{delogblue}\textbf{65.76}
& \textbf{26.59} & \textbf{71.45} & \textbf{64.13} & \cellcolor{delogblue}\textbf{54.06}
& \textbf{21.91} & \textbf{61.30} & \textbf{42.73} & \cellcolor{delogblue}\textbf{38.65} & \cellcolor{proprietarygray}\textbf{52.82}\\

& $\Delta$
& \up{6.31} & \up{8.61} & \up{18.09} & \cellcolor{delogblue}\up{11.00}
& \up{5.17} & \up{8.14} & \up{29.06} & \cellcolor{delogblue}\up{15.21}
& \up{8.17} & \up{0.51} & \up{14.78} & \cellcolor{delogblue}\up{6.25} & \cellcolor{proprietarygray}\up{10.82}
\\

\hline

\multirow{7}{*}{Qwen-2.5-1.5B}
& ICL\(^\dagger\) 
& 50.55 & 41.74 & 6.69 & \cellcolor{delogblue}32.99
& 30.67 & 26.09 & 6.89 & \cellcolor{delogblue}21.22
& 20.56 & 17.78 & 4.99 & \cellcolor{delogblue}14.44 & \cellcolor{proprietarygray}22.88\\

& PostCite\(^\dagger\) 
& 16.36 & 52.46 & 15.40 & \cellcolor{delogblue}28.07
& 22.22 & 48.66 & 16.92 & \cellcolor{delogblue}29.27
& 15.63 & 26.71 & 5.17 & \cellcolor{delogblue}15.84 & \cellcolor{proprietarygray}24.39\\

& PostAttr\(^\dagger\) 
& 16.36 & 52.46 & 4.45 & \cellcolor{delogblue}24.42
& 22.22 & 48.66 & 13.15 & \cellcolor{delogblue}28.01
& 15.63 & 26.71 & 0.62 & \cellcolor{delogblue}14.32 & \cellcolor{proprietarygray}22.25\\

& FRONT\(^\dagger\) 
& 57.74 & 41.36 & 55.70 & \cellcolor{delogblue}51.60
& 29.15 & 24.60 & 50.22 & \cellcolor{delogblue}34.66
& 19.57 & 17.29 & 37.70 & \cellcolor{delogblue}24.85 & \cellcolor{proprietarygray}37.04\\

& T-ALIGN\(^\dagger\) 
& 52.68 & 62.38 & 66.81 & \cellcolor{delogblue}60.62
& 25.06	& 68.38	& 51.44	& \cellcolor{delogblue}48.29
& 19.03 & 57.91 & 31.63 & \cellcolor{delogblue}36.19 & \cellcolor{proprietarygray}48.37\\
\cdashline{2-15}
& \textsc{CAGE} 
& \textbf{59.27} & \textbf{69.81} & \textbf{80.67} & \cellcolor{delogblue}\textbf{69.91}
& \textbf{32.05} & \textbf{71.68} & \textbf{77.34} & \cellcolor{delogblue}\textbf{60.35}
& \textbf{28.52} & \textbf{61.47} & \textbf{43.14} & \cellcolor{delogblue}\textbf{44.38} & \cellcolor{proprietarygray}\textbf{58.21}\\

& $\Delta$
& \up{1.53} & \up{7.43} & \up{13.86} & \cellcolor{delogblue}\up{9.29}
& \up{2.90} & \up{3.30} & \up{25.90} & \cellcolor{delogblue}\up{11.37}
& \up{8.95} & \up{3.56} & \up{5.44} & \cellcolor{delogblue}\up{8.19} & \cellcolor{proprietarygray}\up{9.84}
\\
\hline

\multirow{7}{*}{Qwen-2.5-3B}
& ICL\(^\dagger\) 
& 37.72 & 51.36 & 51.72 & \cellcolor{delogblue}46.93 
& 35.14 & 49.65 & 42.67 & \cellcolor{delogblue}42.49
& 29.12 & 46.31 & 34.34 & \cellcolor{delogblue}36.59 & \cellcolor{proprietarygray}42.00\\

& PostCite\(^\dagger\) 
& 9.58 & 35.30 & 10.94 & \cellcolor{delogblue}18.61 
& 0 & 40.66 & 0 & \cellcolor{delogblue}13.55
& 21.73 & 48.49 & 7.56 & \cellcolor{delogblue}25.93 & \cellcolor{proprietarygray}19.36\\

& PostAttr\(^\dagger\) 
& 9.58 & 35.30 & 36.29 & \cellcolor{delogblue}27.06 
& 0 & 40.66 & 0 & \cellcolor{delogblue}13.55
& 21.73 & 48.49 & 1.31 & \cellcolor{delogblue}23.84 & \cellcolor{proprietarygray}21.48\\

& FRONT\(^\dagger\) 
& 55.15 & 44.01 & 62.72 & \cellcolor{delogblue}53.96 
& 25.67 & 29.86 & 44.48 & \cellcolor{delogblue}33.34
& 18.69 & 25.37 & 37.40 & \cellcolor{delogblue}27.15 & \cellcolor{proprietarygray}38.15\\

& T-ALIGN\(^\dagger\) 
& 55.19 & 63.76 & 78.64 & \cellcolor{delogblue}65.86
& 20.97	& 65.79	& 60.25	& \cellcolor{delogblue}49.0
& 22.52 & \textbf{64.38} & 42.01 & \cellcolor{delogblue}42.97 & \cellcolor{proprietarygray}52.61\\
\cdashline{2-15}
& \textsc{CAGE} 
& \textbf{59.98} & \textbf{70.08} & \textbf{83.09} & \cellcolor{delogblue}\textbf{71.05}
& \textbf{33.68} & \textbf{71.75} & \textbf{78.39} & \cellcolor{delogblue}\textbf{61.27}
& \textbf{30.69} & 61.63 & \textbf{48.55} & \cellcolor{delogblue}\textbf{46.96} & \cellcolor{proprietarygray}\textbf{59.76}\\

& $\Delta$
& \up{4.79} & \up{6.32} & \up{4.45} & \cellcolor{delogblue}\up{5.19}
& \up{0.54} & \up{5.96} & \up{18.14} & \cellcolor{delogblue}\up{12.27}
& \up{1.57} & \down{-2.75} & \up{6.54} & \cellcolor{delogblue}\up{3.99} & \cellcolor{proprietarygray}\up{7.15}
\\

\hline

\multirow{7}{*}{Qwen-2.5-7B}
& ICL\(^\dagger\) 
& 58.94 & 54.34 & 75.46 & \cellcolor{delogblue}62.91 
& \textbf{36.33} & 42.28 & 56.09 & \cellcolor{delogblue}44.9
& 28.27 & 37.13 & 44.13 & \cellcolor{delogblue}36.51 & \cellcolor{proprietarygray}48.11\\

& PostCite\(^\dagger\) 
& 27.52 & 45.93 & 4.19 & \cellcolor{delogblue}25.88 
& 25.58 & 54.9 & 13.77 & \cellcolor{delogblue}31.42
& 21.82 & 22.23 & 7.03 & \cellcolor{delogblue}17.03 & \cellcolor{proprietarygray}24.78\\

& PostAttr\(^\dagger\)
& 27.52 & 45.93 & 17.92 & \cellcolor{delogblue}30.46 
& 25.58 & 54.9 & 12.46 & \cellcolor{delogblue}30.98
& 21.82 & 22.23 & 0.96 & \cellcolor{delogblue}15.00 & \cellcolor{proprietarygray}25.48\\

& FRONT\(^\dagger\) 
& \textbf{64.58} & 60.08 & 58.27 & \cellcolor{delogblue}60.98 
& 32.41 & 55.56 & 67.35 & \cellcolor{delogblue}51.77
& 28.27 & 54.14 & 56.61 & \cellcolor{delogblue}46.34  & \cellcolor{proprietarygray}53.03\\

& T-ALIGN\(^\dagger\) 
& 55.04 & 66.22 & 83.57 & \cellcolor{delogblue}68.28
& 25.57	& 69.16	& 62.7	& \cellcolor{delogblue}52.48
& 24.30 & \textbf{63.79} & 47.02 & \cellcolor{delogblue}45.04 & \cellcolor{proprietarygray}55.27\\
\cdashline{2-15}
& \textsc{CAGE} 
& 60.98 & \textbf{70.09} & \textbf{85.32} & \cellcolor{delogblue}\textbf{72.12}
& 33.14 & \textbf{71.61} & \textbf{80.04} & \cellcolor{delogblue}\textbf{61.60}
& \textbf{31.32} & 61.31 & \textbf{53.44} & \cellcolor{delogblue}\textbf{48.70} & \cellcolor{proprietarygray}\textbf{60.81}\\

& $\Delta$
& \down{-3.60}& \up{3.87}& \up{1.75}& \cellcolor{delogblue}\up{3.84}
& \down{-3.19}& \up{2.45}& \up{12.69}& \cellcolor{delogblue}\up{9.12}
& \up{3.05}& \down{-2.48}& \up{6.42}& \cellcolor{delogblue}\up{2.36}& \cellcolor{proprietarygray}\up{5.54}
\\
\hline
\multirow{1}{*}{Qwen3.5-9B}
& \textsc{CAGE} 
& 60.21 & 70.08 & 87.87 & \cellcolor{delogblue}\textbf{72.72}
& 33.74 & 71.70 & 80.95 & \cellcolor{delogblue}\textbf{62.13}
& 29.83 & 61.55 & 57.79 & \cellcolor{delogblue}\textbf{49.72} & \cellcolor{proprietarygray}\textbf{61.52} \\

\hline
\end{tabular}
\caption{Main results on ASQA, ExpertQA, and ELI5. Qwen3.5-9B is fixed as CMI and Qwen-2.5 models serve as SCR, except the last row, which uses separate Qwen3.5-9B models for both stages. The answerable setting and $\dagger$ results follow \cite{song2025measuring}. $\Delta$ is the gain over the strongest baseline; \textbf{Average} is mean \textbf{TRUST}. LLaMA results are in the Appendix.}
\label{main_results_qwen}
\end{table*}

\subsubsection{Cognitive Attribution Map Induction Model}
\label{sec:cmi}

CMI implements the first stage of CAGE by learning to generate the textual
encoding of a cognitive attribution map. Given a query \(Q\) and its top-\(k\)
retrieved documents \(\mathcal{D}_k\), CMI predicts
\begin{equation}
    \hat{z}_{\mathrm{map}}
=
f_{\theta_g}(Q,\mathcal{D}_k,\mathcal{I}_\mathrm{map}).
\end{equation}
Here, \(\hat{z}_{\mathrm{map}}\) denotes the generated textual encoding of the
cognitive attribution map, \(\mathcal{I}_\mathrm{map}\) is a task specific instruction for CMI. When well formed, it specifies an induced
graph-structured cognitive attribution map \(\hat{\mathbf{G}}\).
 If the generated encoding
corresponds to an empty map, CMI indicates that no answer unit can be supported
by the retrieved documents. Otherwise, the encoded map specifies answer-centered support subgraphs following
the schema in Eq.~\ref{node_schema}, including answer nodes, supporting document
nodes, support edges, and descriptions.

Thus, CMI learns to decide what can be supported, abstract supported answer
units, and assign document evidence to these units before any final answer
sentence is decoded. The graph-structured map remains the intermediate
reasoning object, while its textual encoding is the sequence generated and
optimized by the LLM.

CMI is trained to generate the target map encoding \(z_{\mathrm{map}}^{*}\)
constructed by VCAS in Eq.~\ref{eq:VCAS-graph-target}. For non-empty maps, this encoding preserves the
typed nodes, support edges, and support descriptions of \(\mathbf{G}^{*}\). For
empty maps, it produces a schema-preserving empty-map encoding that contains no
evidence-supported answer nodes, supporting document nodes, or support edges,
but records why the retrieved documents fail to support the question.

Let \(X_g=(Q,\mathcal{D}_k,\mathcal{I}_\mathrm{map})\), CMI is optimized by autoregressive maximum likelihood:
\begin{equation}
\begin{aligned}
    \mathcal{L}_{\mathrm{CMI}}
    =
    -
    \mathbb{E}_{\mathcal{S}_{\mathrm{CMI}}}
    \sum_{t=1}^{|z_{\mathrm{map}}^{*}|}
    \log P_{\theta_g}
    \left(
        z^{*}_{\mathrm{map},t}
        \mid
        z^{*}_{\mathrm{map},<t}, X_g
    \right).
\end{aligned}
\label{eq:cmi-objective}
\end{equation}
where the expectation is taken over \((X_g,z_\mathrm{map}^{*})\sim\mathcal{S}_{\mathrm{CMI}}\).

At inference time, CMI generates the map encoding \(\hat{z}_{\mathrm{map}}\) and
passes it to SCR. If the encoding specifies an empty map, it preserves the attribution-map
schema and records why the retrieved documents fail to support the question,
but contains no concrete evidence-supported answer nodes, supporting document
nodes, or support edges. Otherwise, it provides the encoded support subgraphs,
retained document content, and map-specified citation targets for citation
reasoning.

\subsubsection{Structured Citation Reasoning Model}
\label{sec:scr}

SCR implements the second stage of CAGE by reasoning over the encoded cognitive
attribution map and realizing the underlying map as a citation-grounded
long-form answer. Given the \(Q\) and 
\(z_{\mathrm{map}}\), SCR predicts the final answer:
\begin{equation}
    \hat{\mathcal{Y}}
    =
    f_{\theta_a}
    \left(
        Q,
        z_{\mathrm{map},},\mathcal{I}_\mathrm{ans}
    \right).
\end{equation}
Here, \(\mathcal{I}_\mathrm{ans}\) is an instruction for SCR and \(\hat{\mathcal{Y}}=\{(\hat{s}_j,\hat{c}_j)\}_{j=1}^{m}\),
where \(\hat{s}_j\) is a generated claim and \(\hat{c}_j\) is its citation set.

SCR performs three map-aligned operations: realizing answer nodes as
sentence-level claims, aligning each claim with citation IDs specified by the
support-subgraph document assignments, and organizing the cited claims into a
coherent long-form answer. Since the output order need not follow
the support-subgraph order, correctness is defined by a permutation \(\pi\) satisfying
\begin{equation}
    s_{\pi(r)}
    \equiv
    a_r,
    \quad
    c_{\pi(r)}
    =
    \alpha_{\mathbf{G}}(a_r),
    \quad
    \forall r\in\{1,\ldots,m\},
    \label{eq:scr-claim-citation-alignment}
\end{equation}
where \(\alpha_{\mathbf{G}}(a_r)\) denotes the citation identifiers of the
document nodes assigned to answer node \(a_r\) in \(\mathbf{G}\).

For autoregressive training, the map-aligned answer itself serves as the target
sequence, written as claim--citation units:
\begin{equation}
    \mathcal{Y}
    =
    \oplus_{j=1}^{m}
    [s_j;c_j].
\end{equation}
Let \(X_a=(Q,z_{\mathrm{map}}, \mathcal{I}_\mathrm{ans})\), SCR is optimized by
autoregressive maximum likelihood over the final answer sequence:
\begin{equation}
\mathcal{L}_{\mathrm{SCR}}
=
-
\mathbb{E}_{(X_a,\mathcal{Y})\sim\mathcal{S}_{\mathrm{SCR}}}
\sum_{t=1}^{|\mathcal{Y}|}
\log
P_{\theta_a}
\left(
    \mathcal{Y}_{t}
    \mid
    \mathcal{Y}_{<t},
    X_a
\right).
\label{eq:scr-objective}
\end{equation}
This objective trains SCR to generate cited claims aligned with the
map-specified document IDs encoded in \(z_{\mathrm{map}}\).

\section{Experiments}
\label{sec:experiments}

\subsection{Experimental settings}

\textbf{Datasets.} We evaluate our method on three representative long-form QA benchmarks for inline citation generation: ASQA \cite{stelmakh-etal-2022-asqa}, ELI5
\cite{fan-etal-2019-eli5}, and ExpertQA \cite{malaviya-etal-2024-ExpertQA}.

\textbf{Baselines.} We compare \textsc{CAGE} with six representative baselines: \textit{ICL}, \textit{PostCite}, \textit{PostAttr} \cite{gao-etal-2023-enabling}, \textit{FRONT} \cite{huang2024learning}, \textit{T-ALIGN} \cite{song2025measuring}, and \textit{Ground-GRPO} \cite{sim2025lessonstraininggroundedllms}, covering in-context generation, post-hoc attribution, preference learning, grounded answering, and GRPO-based alignment. 
We exclude START~\cite{huang2024advancing}, LongCite~\cite{zhang2025longcite}, and SelfCite~\cite{chuang2025selfcite} due to differences in reproducibility, task formulation, and evaluation protocol.

\textbf{LLMs.} We evaluate CAGE with \texttt{Qwen-2.5}, \texttt{Qwen-3} series and \texttt{Qwen-3.5} as SCR models. In CMI, we use Qwen3.5-9B by default. We further compare CAGE with \texttt{GPT-5.5} and \texttt{Claude-Sonnet-4.6}. The implementation details can be found in Appendix.

\textbf{Metrics.}
We adopt TRUST~\cite{song2025measuring} as the overall metric, which averages answer calibration \(\mathrm{EM}^{F1}_{\mathrm{AC}}\), refusal behavior \(\mathrm{F1}_{\mathrm{RG}}\), and citation groundedness \(\mathrm{F1}_{\mathrm{CG}}\). Here, \(\mathrm{EM}^{F1}_{\mathrm{AC}}\) measures correctness on answerable questions while accounting for the model's decision to answer, \(\mathrm{F1}_{\mathrm{RG}}\) evaluates answering/refusal decisions over answerable and unanswerable questions, and \(\mathrm{F1}_{\mathrm{CG}}\) measures citation precision and recall using the TRUE model following prior work~\cite{gao-etal-2023-enabling,song2025measuring}. The final score is:
\begin{equation}
    \mathbf{TRUST}
    =
    \frac{1}{3}
    \left(
        \mathrm{EM}^{F1}_{\mathrm{AC}}
        +
        \mathrm{F1}_{\mathrm{RG}}
        +
        \mathrm{F1}_{\mathrm{CG}}
    \right).
\end{equation}
Here, the details of metric can be found in the Appendix.

\subsection{Experimental Results and Analysis}

\subsubsection{Main Results} The experimental results are displayed in Table \ref{main_results_qwen} and the experimental analyses are listed as follows:

(1) CAGE consistently achieves the best TRUST scores across Qwen-2.5 SCR models of different scales, indicating that its effectiveness is not tied to a specific model size. Averaged across datasets, CAGE improves TRUST over the strongest baseline by 10.82\%, 9.84\%, 7.15\%, and 5.54\% with Qwen-2.5-0.5B, 1.5B, 3B, and 7B, respectively. The gains are particularly pronounced in attribution grounding: on ExpertQA, CAGE improves F1$_{\mathrm{GC}}$ by up to 29.06\%. These results support our theoretical analysis that graph-specified evidence paths contract the attribution space, reducing attribution ambiguity.

(2) Across the three TRUST dimensions, CAGE consistently improves answer calibration, refusal behavior, and citation grounding across different model--dataset settings. The gains are especially notable in citation grounding, with F1$_{\mathrm{GC}}$ reaching 70.5\% on ASQA with Qwen-2.5-0.5B and 78.4\% on ExpertQA with Qwen-2.5-3B, while maintaining strong performance on the other dimensions. These results suggest that the graph-guided intermediate representation organizes evidence before generation, enabling more accurate claim-level attribution without compromising overall response quality.

\begin{figure}[t]
    \centering
    \includegraphics[width=1\linewidth]{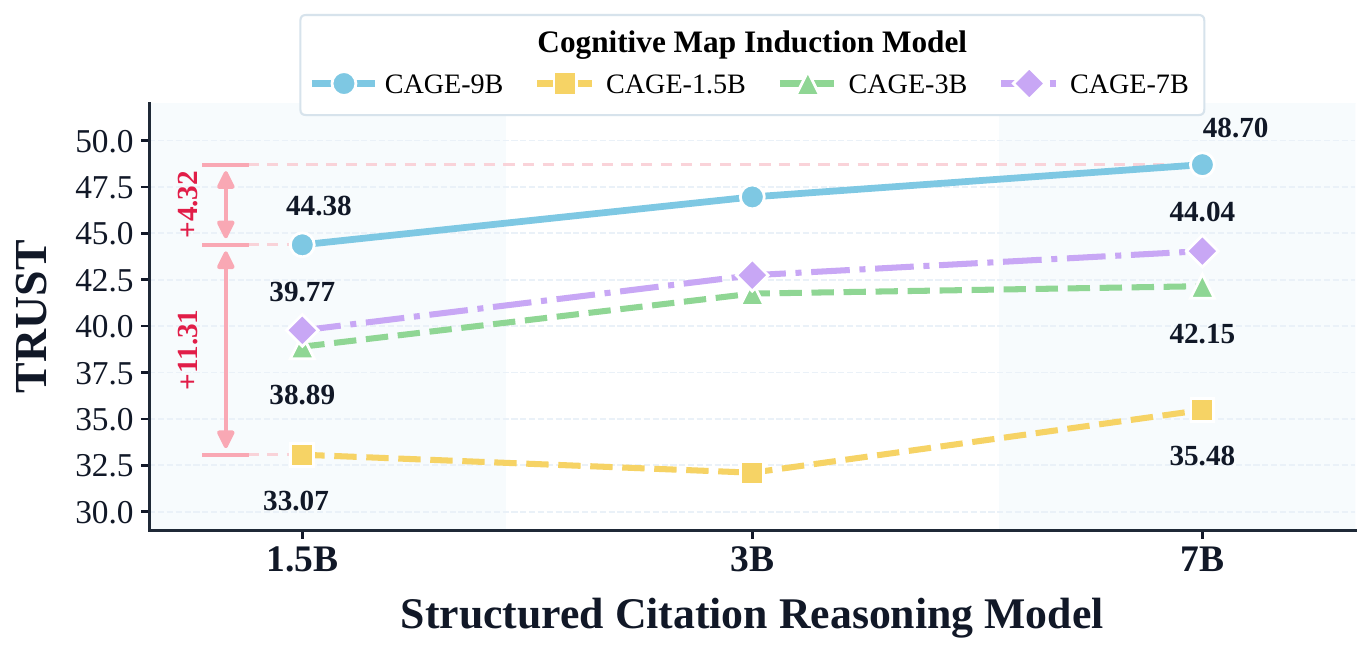}
    \caption{Effect of model scale on TRUST performance. CAGE-9B uses Qwen3.5-9B in CMI stage, while CAGE-1.5B/3B/7B use Qwen-2.5 models of the corresponding sizes for map induction. The SCR model is varied from 1.5B to 7B, showing that scaling map induction brings larger gains than merely scaling SCR model.
    }
    \label{fig:trust_three_sizes_three_datasets}
\end{figure}

(3) Finally, Table~\ref{main_results_qwen} shows that scaling the SCR model yields only marginal gains once a high-quality graph is available. Qwen3.5-9B achieves the highest average TRUST score of 61.52\%, but improves over the Qwen-2.5-7B SCR model by only 0.71\%. This suggests that reliable support graphs, rather than model scaling, are the primary driver of evidence alignment and attribution quality.

\subsubsection{Effect of Model Scale in CAGE}

As shown in Figure~\ref{fig:trust_three_sizes_three_datasets}, scaling the SCR model improves TRUST, but increasing map induction capacity yields larger gains. With CAGE-9B as the induction model, scaling the SCR model from 1.5B to 7B improves TRUST 4.32\%. In contrast, fixing the SCR model and replacing the same-size induction model with CAGE-9B brings larger gains: 33.07\% $\rightarrow$ 44.38\% on 1.5B (+11.31\%), 41.75\% $\rightarrow$ 46.96\% on 3B (+5.21\%), and 44.04\% $\rightarrow$ 48.70\% on 7B (+4.66\%). These results indicate that TRUST is more constrained by map induction quality than by structured citation reasoning capacity, since stronger induction better establishes evidence--answer alignments and contracts the attribution space before generation.

\subsubsection{Ablation Study}

\begin{figure}[t]
    \centering
    \includegraphics[width=1\linewidth]{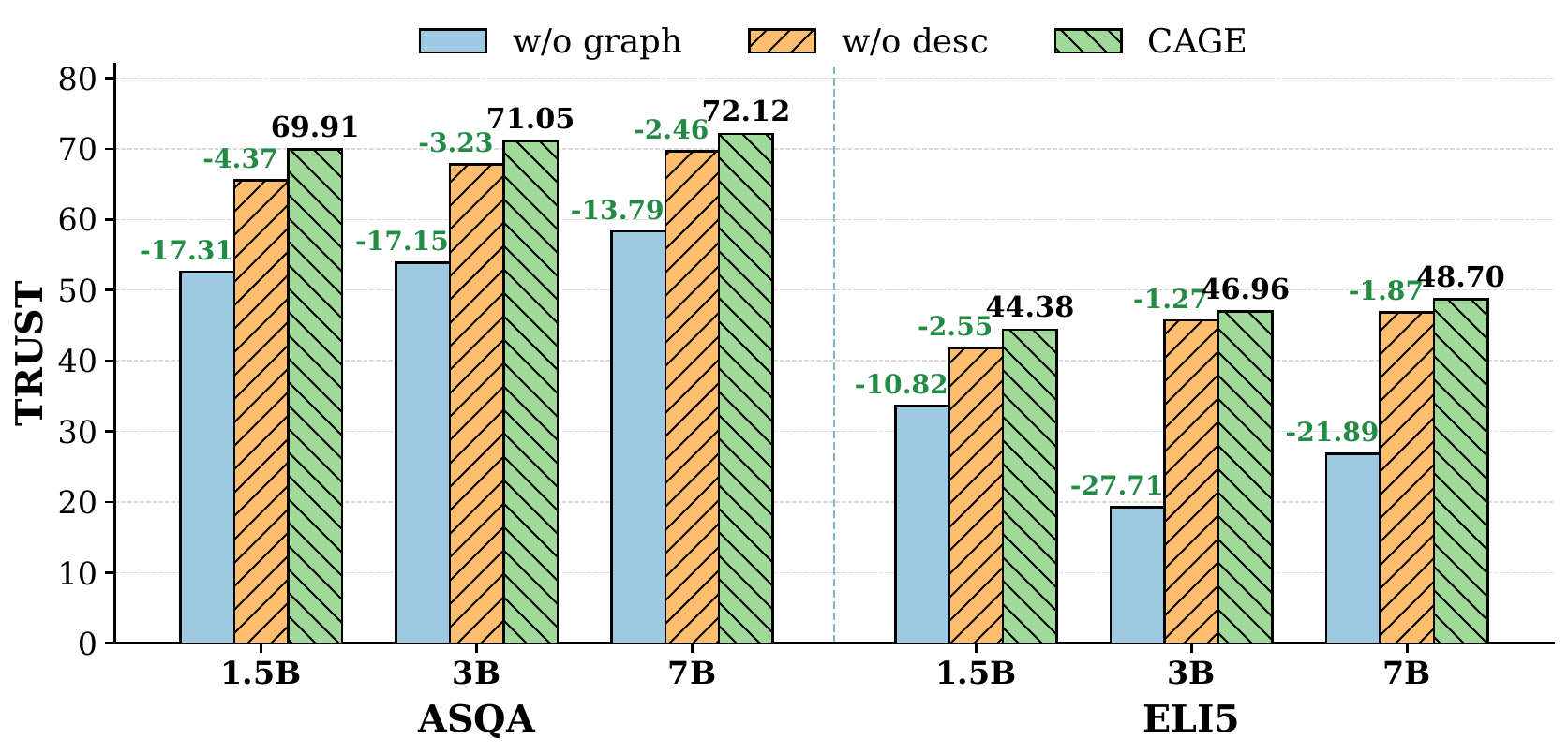}
    \caption{Ablation study. We fix the structured citation reasoning model size (1.5B/3B/7B) and ablate map induction model, including ``\textit{w/o graph}'' (no structure, documents retained) and ``\textit{w/o desc}'' (no graph descriptions).}
    \label{fig:ablation111}
\end{figure}

We conduct ablation experiments for the CMI stage, as reported in Figure~\ref{fig:ablation111}. Specifically, \textit{w/o graph} removes the support graph while retaining the documents mentioned in the subgraphs, and \textit{w/o desc} removes the textual descriptions associated in subgraph.

(1) In the \textit{w/o graph} setting, TRUST drops substantially despite retaining the same documents; for example, on ASQA with Qwen-2.5-1.5B, it decreases from 69.91\% to 52.60\%. This shows that CAGE benefits from explicit graph structure rather than document filtering alone. Without the graph, the generator must infer claim--document assignments from an unstructured document set, whereas the support graph explicitly constrains these assignments, supporting our attribution-space contraction analysis.

(2) In the \textit{w/o desc} setting, performance also declines, but less severely than removing the graph structure; for example, TRUST on ASQA with Qwen-2.5-1.5B drops from 69.91\% to 65.54\%. This suggests that textual descriptions help the generator interpret the semantic role of each support unit, while the main attribution-space contraction still comes from the graph structure. 

\subsubsection{Analysis of CMI under Different SCR Backbones}

We analyze CAGE from three perspectives: its effectiveness over strong end-to-end baselines, the plug-and-play generality of CMI across SCR backbones, and its ability to reduce evidence-boundary overrun, as reported in Tables~\ref{tab:qwen_results1111} and~\ref{tab:asqa_ebo}.

(1) \textit{CAGE outperforms strong end-to-end baselines, highlighting the benefit of attribution-space contraction.} On ASQA, CAGE with Claude-4.6 achieves 76.62\% TRUST, outperforming end-to-end Claude-4.6 by 7.21\%. With Qwen3-4B, it also exceeds Ground-GRPO by 6.94\%. These gains indicate that stronger backbones or preference optimization alone cannot fully resolve citation attribution. By inducing a Cognitive Map before citation reasoning, CAGE narrows the claim--document assignment space and reduces unsupported attributions.

(2) \textit{CMI offers practical plug-and-play utility across heterogeneous SCR backbones.} Its consistent gains with both Claude and GPT models show that CMI can be directly integrated into proprietary LLM systems without additional training. For example, on ExpertQA, CAGE with Claude-4.6 improves TRUST from 55.33\% to 66.43\%. CAGE therefore provides a reusable attribution service that improves generators across model families and scales without generator-specific retraining or architectural modifications.

\begin{table}[t]
  \centering
  \footnotesize
  \setlength{\tabcolsep}{5pt}
  \renewcommand{\arraystretch}{1}
  \begin{tabular}{lcccc}
    \toprule
    Method &
    $\mathrm{EM}^{\mathrm{F1}}_{\mathrm{AC}}$ &
    $\mathrm{F1}_{\mathrm{RG}}$ &
    $\mathrm{F1}_{\mathrm{CG}}$ &
    \textbf{TRUST} \\
    \midrule

    \multicolumn{5}{c}{ASQA} \\
    \midrule

    \rowcolor{proprietarygray}
    Ground-GRPO$_{\text{4B}}$
      & 47.22 & 62.88 & 81.94 & 64.01 \\
    CMI$_{\text{9B}}$+SCR$_{\text{Qwen3-4B}}$
      & 59.50 & 66.54 & 86.82 & 70.95\textcolor{red}{$_{6.94\uparrow}$} \\

    \rowcolor{proprietarygray}
    GPT-5.5
      & 66.18 & 61.54 & 70.69 & 66.14 \\

    CMI$_{\text{9B}}$+SCR$_{\text{GPT-5.5}}$
      & 61.41 & 70.09 & 92.32
      & 74.61\textcolor{red}{$_{8.47\uparrow}$} \\

    \rowcolor{proprietarygray}
    Claude-4.6
      & 70.46 & 64.56 & 73.22 & 69.41 \\
    
    CMI$_{\text{9B}}$+SCR$_{\text{Claude-4.6}}$
      & 65.29 & 70.09 & 94.48
      & 76.62\textcolor{red}{$_{7.21\uparrow}$} \\

    \midrule
    \multicolumn{5}{c}{ExpertQA} \\
    \midrule

    \rowcolor{proprietarygray}
    Ground-GRPO$_{\text{4B}}$
      & 20.56 & 67.14 & 59.37 & 49.02 \\

    CMI$_{\text{9B}}$+SCR$_{\text{Qwen3-4B}}$
      & 36.27 & 62.22 & 78.81 & 62.41\textcolor{red}{$_{13.39\uparrow}$} \\

    \rowcolor{proprietarygray}
    GPT-5.5
      & 46.74 & 54.46 & 49.78 & 50.33 \\

     CMI$_{\text{9B}}$+SCR$_{\text{GPT-5.5}}$
      & 32.88 & 71.74 & 89.15
      &64.59\textcolor{red}{$_{14.26\uparrow}$} \\

    \rowcolor{proprietarygray}
    Claude-4.6
      & 47.66 & 64.54 & 53.80 & 55.33 \\

    CMI$_{\text{9B}}$+SCR$_{\text{Claude-4.6}}$
      & 36.70 & 71.74 & 90.86
      & 66.43\textcolor{red}{$_{11.10\uparrow}$} \\

    \bottomrule
  \end{tabular}

  \caption{Backbone capability and proprietary model comparison. 9B denotes Qwen3.5-9B. We additionally finetune Qwen3-4B for the SCR stage to enable a direct comparison with Ground-GRPO.}

  \label{tab:qwen_results1111}
\end{table}

(3) \textit{CAGE substantially reduces evidence-boundary overrun.}
Following the atomic-fact decomposition paradigm of FActScore~\citep{min-etal-2023-factscore}, we use GPT-5.5 to decompose each valid long-form answer into atomic facts and apply TRUE to assess their support against the top-5 retrieved documents considered jointly. Unsupported facts are treated as evidence-boundary overruns. We report their overall proportion (Atomic-EBO), mean per-answer proportion (EBO-Severity), and occurrence rate across answers (Response-EBO), with formal definitions provided in the Appendix.

As shown in Table~\ref{tab:asqa_ebo}, direct GPT-5.5 and Claude-4.6 yield Atomic-EBO scores of 38.00\% and 40.85\%, which fall to 7.21\% and 5.77\% after incorporating CMI with the corresponding SCR backbones. Similar reductions across EBO-Severity and Response-EBO show that stronger generation alone cannot ensure evidence-bounded answers. By constraining answer units to supporting evidence through a Cognitive Map, CAGE substantially reduces overrun at both the fact and answer levels.

\section{Related Work}

Existing methods for inline citation generation can be categorized into
parameter-updating and prompt-based approaches. \textit{Parameter-updating}
methods train models on citation-oriented supervision. Early work constructs
statement--document attribution data with NLI models or advanced LLMs and
improves citation quality through supervised fine-tuning or preference
optimization~\citep{ye2024effective,huang2024learning,li2024improvingattributedtextgeneration}.
Later studies explore self-improving data synthesis
~\citep{huang2024advancing}, coarse-to-fine construction for long-context
generation~\citep{zhang2025longcite}, context-ablation-based preference
optimization~\citep{chuang2025selfcite}, and grounding from parametric or
external knowledge~\citep{khalifasource,shen2025transparentize}.

\textit{Prompt-based} methods keep the backbone LLM frozen and use prompting or
decomposition to elicit citation behavior. ALCE~\citep{gao-etal-2023-enabling}
prompts LLMs with retrieved documents, while later work adopts
chain-of-thought, search-based expansion, or query/document selection
~\citep{Ji_Liu_Du_Ng_2024,li2025think,hirsch2025laquer}.
Attribute First, then Generate~\citep{slobodkin2024attribute} organizes selected evidence spans into sentence-level plans before generation. In contrast, CAGE models attribution as an explicit intermediate reasoning structure by inducing validated answer--document support graphs, enabling generation to follow structured attribution constraints rather than evidence planning alone.

\begin{table}[t]
  \centering
  \footnotesize
  \setlength{\tabcolsep}{8.5pt}
  \renewcommand{\arraystretch}{1}
  \begin{tabular}{lccc}
    \toprule
    Method &
    A-EBO$\downarrow$ &
    S-EBO$\downarrow$ &
    R-EBO$\downarrow$ \\
    \midrule
    GPT-5.5
      & 38.00 & 34.33 & 62.71 \\
    Claude-4.6
      & 40.85 & 36.90 & 83.48 \\
    CMI$_{\text{9B}}+$SCR$_{\text{Qwen-7B}}$
      & 14.69 & 13.11 & 39.28 \\
    CMI$_{\text{9B}}+$SCR$_{\text{GPT-5.5}}$
      & \underline{7.21} & \underline{7.30} & \textbf{15.49} \\
    CMI$_{\text{9B}}+$SCR$_{\text{Claude-4.6}}$
      & \textbf{5.77} & \textbf{5.58} & \underline{18.94} \\
    \bottomrule
  \end{tabular}
  \caption{Evidence-boundary overrun on ASQA. A-EBO (Atomic-EBO) measures the proportion of unsupported atomic facts, S-EBO (EBO-Severity) their mean per-answer proportion, and R-EBO (Response-EBO) the proportion of answers containing any unsupported fact. Lower is better.}
  \label{tab:asqa_ebo}
\end{table}

\section{Conclusion and Limitations}

We propose \textsc{CAGE}, a two-stage framework that introduces a cognitive attribution map for citation-grounded long-form generation. By modeling answer--evidence attribution before generation, \textsc{CAGE} yields more reliable claim-level citations, clearer evidential boundaries, and less attribution ambiguity. Experiments on ASQA, ELI5, and ExpertQA show consistent gains across diverse open-source and proprietary LLMs. As a reusable attribution service, \textsc{CAGE} improves diverse
generation backbones without generator-specific retraining or
architectural modifications. These results demonstrate both the effectiveness and practical deployability of explicit attribution modeling across model families and scales.

The two-stage design inevitably introduces error propagation: the map induction module fails to preserve supporting evidence for 15.42\% of instances on average, limiting downstream recovery. Nevertheless, CAGE consistently improves citation-grounded generation across datasets and backbones, while making attribution errors more identifiable and measurable. Future work will focus on improving attribution map induction and reducing cross-stage error propagation.

\bibliography{aaai2027}

\newpage

\clearpage

\appendix

\section{Implementation details.}
For the Valid Cognitive Attribution Synthesis stage, we use GPT-5.5 as both the validator and generator. We implement the training of the Cognitive Attribution Map Induction Model and Structured Citation Reasoning Model using LLaMA-Factory. Both components are optimized via full-parameter supervised fine-tuning with DeepSpeed ZeRO-3 and BF16 precision on 4 NVIDIA A800-80G GPUs. The graph induction model uses a per-device batch size of 6, while the citation reasoning model uses a per-device batch size of 8 with gradient accumulation of 2. We train both models for three epochs using AdamW with a learning rate of \(1\times10^{-5}\), cosine learning-rate scheduling, and a warmup ratio of 0.1. The random seed is set to 42, and weight decay is set to zero for all experiments.

\begin{table*}[t]
\centering
\footnotesize
\setlength{\tabcolsep}{3pt} 
\begin{tabular}{
cc
cccc
cccc
cccc
c
}
\hline
\multirow{3}{*}{Size} & \multirow{3}{*}{Type} 
& \multicolumn{4}{c}{\makecell{ASQA(610 answerable, \\338 unanswerable)}}
& \multicolumn{4}{c}{\makecell{EXPERTQA(682 answerable,\\1487 unanswerable)}} 
& \multicolumn{4}{c}{\makecell{ELI5(207 answerable,\\793 unanswerable)}}
& \multirow{3}{*}{Average}  \\
\cline{3-14}

& & \multicolumn{2}{c}{Truthfulness} & \multicolumn{1}{c}{AG} & \multirow{2}{*}{Trust} 
  & \multicolumn{2}{c}{Truthfulness} & \multicolumn{1}{c}{AG} & \multirow{2}{*}{Trust} 
  & \multicolumn{2}{c}{Truthfulness} & \multicolumn{1}{c}{AG} & \multirow{2}{*}{Trust} \\
\cline{3-5} \cline{7-9} \cline{11-13}

& & F1$_{AC}$ & F1$_{GR}$ & F1$_{GC}$ & 
  & F1$_{AC}$ & F1$_{GR}$ & F1$_{GC}$ & 
  & F1$_{AC}$ & F1$_{GR}$ & F1$_{GC}$ & \\
  
\hline

\multirow{7}{*}{2-7b}
& ICL 
& 0.00 & 26.28 & 0.00 & 8.76
& 0.00 & 41.01 & 9.25 & 16.84
& 0.50 & 46.71 & 5.04 & 15.57 & 13.72\\

& PostCite 
& 0.07 & 35.23 & 0.00 & 11.77
& 4.85 & 44.27 & 5.23 & 18.12
& 1.86 & 44.98 & 13.80 & 17.29 & 15.73\\

& PostAttr 
& 0.07 & 35.23 & 0.00 & 11.77
& 4.85 & 44.27 & 2.26 & 17.13
& 1.86 & 44.98 & 0.00 & 15.61 & 14.84\\

& FRONT 
& \textbf{60.47} & 39.15 & 68.86 & 56.16
& 9.33 & 23.92 & 74.75 & 36.00
& 21.66 & 17.15 & \textbf{52.72} & 30.51 & 40.89\\

& T-ALIGN 
& 52.48 & 66.12 & \textbf{83.94} & 67.51 
& 25.03 & 67.91 & 62.46 & 51.8
& 22.54 & \textbf{63.27} & 47.35 & 44.39 & 54.57\\
\cdashline{2-15}

& \textsc{CAGE} 
& 58.59 & \textbf{69.62} & 83.34 & \textbf{70.52}
& \textbf{31.25} & \textbf{71.24} & \textbf{75.19} & \textbf{59.22}
& \textbf{27.61} & 61.44 & \underline{49.15} & \textbf{46.07} & \textbf{58.60}\\

& $\Delta$
& \down{1.88}& \up{3.50}& \down{0.60}& \up{3.01}
& \up{6.22}& \up{3.33}& \up{0.44}& \up{7.42}
& \up{5.07}& \down{1.83}& \down{3.57}& \up{1.68} & \up{4.04}
\\

\hline

\multirow{7}{*}{3.2-1b}
& ICL 
& 35.95 & 50.94 & 9.96 & 32.28
& 21.55 & 32.83 & 9.04 & 21.14
& 12.87 & 27.10 & 5.23 & 15.07 & 22.83\\

& PostCite 
& 0.59 & 50.22 & 0.24 & 17.02
& 5.48 & 49.1 & 2.67 & 19.08
& 2.04 & 50.88 & 1.02 & 17.98 & 18.03\\

& PostAttr 
& 0.48 & 48.42 & 0.00 & 16.30
& 8.24 & 47.72 & 1.5 & 19.15
& 2.04 & 50.88 & 0.07 & 17.66 & 17.70\\

& FRONT 
& 48.22 & 54.48 & 48.29 & 50.33
& 20.83 & 29.26 & 37.45 & 29.18
& 16.11 & 20.76 & 30.19 & 22.35 & 33.95\\

& T-ALIGN 
& 38.64 & 58.61 & \textbf{79.35} & 58.87
& 20.32 & 64.87 & 62.1 & 49.1
& 13.20 & 59.35 & \textbf{48.21} & 40.25 & 49.41\\
\cdashline{2-15}

& \textsc{CAGE} 
& \textbf{60.20} & \textbf{70.07} & 77.19 & \textbf{69.16}
& \textbf{31.17} & \textbf{71.85} & \textbf{74.26} & \textbf{50.09}
& \textbf{27.69} & \textbf{61.46} & 41.25 & \textbf{43.47} & \textbf{54.24}\\

& $\Delta$
& \up{11.98}& \up{11.46}& \down{2.16}& \up{10.29}
& \up{10.34}& \up{6.98}& \up{12.16}& \up{0.99}
& \up{14.49}& \up{2.11}& \down{6.96}& \up{3.22} & \up{4.83}
\\

\hline

\multirow{7}{*}{3.2-3b}
& ICL 
& 2.04 & 27.98 & 53.95 & 27.99
& 33.5 & 51.21 & 38.37 & 41.03
& 18.55 & 55.56 & 30.70 & 34.94 & 34.65\\

& PostCite 
& 31.03 & 56.59 & 2.29 & 36.87
& 25.68 & 38.11 & 5.29 & 23.03
& 18.12 & 25.14 & 4.44 & 15.90 & 25.27\\

& PostAttr 
& 29.76 & 56.71 & 4.69 & 30.39
& 25.45 & 38.58 & 3.4 & 22.48
& 18.48 & 25.14 & 0.53 & 14.72 & 22.53\\

& FRONT 
& \textbf{63.19} & 49.45 & 57.46 & 56.70
& 27.24 & 43.34 & 50.91 & 40.5
& 19.95 & 32.21 & 41.97 & 31.38 & 42.86\\

& T-ALIGN 
& 59.82 & 66.38 & 84.21 & 70.14
& 11.72 & 56.93 & 78.35 & 49.0
& 18.33 & \textbf{62.79} & \textbf{55.87} & 45.66 & 54.93\\
\cdashline{2-15}

& \textsc{CAGE} 
& 60.45 & \textbf{69.95} & \textbf{87.65} & \textbf{72.68}
& \textbf{33.60} & \textbf{71.61} & \textbf{80.83} & \textbf{62.02}
& \textbf{30.70} & 61.39 & 51.92 & \textbf{48.00} & \textbf{60.90}\\

& $\Delta$
& \down{2.74}& \up{3.57}& \up{3.44}& \up{2.54}
& \up{6.36}& \up{14.68}& \up{2.48}& \up{13.02}
& \up{12.37}& \down{1.40}& \down{3.95}& \up{2.34}& \up{5.97}
\\

\hline

\multirow{7}{*}{3-8b}
& ICL 
& 3.01 & 28.58 & 86.50 & 39.36
& 2.82 & 42.5 & 69.46 & 38.26
& 0.00 & 44.23 & 0.00 & 14.74 & 30.79\\

& PostCite 
& 32.98 & 53.31 & 28.01 & 38.10
& 14.06 & 50.08 & 7.09 & 23.74
& 20.80 & 45.88 & 8.06 & 24.91 & 28.92\\

& PostAttr 
& 32.98 & 53.31 & 5.95 & 30.75
& 14.06 & 50.08 & 6.29 & 23.47
& 20.80 & 45.88 & 1.25 & 22.64 & 25.62\\

& FRONT 
& \textbf{62.25} & 41.62 & 66.14 & 56.67
& 30.34 & 24.92 & 56.7 & 37.32
& 18.99 & 17.85 & 44.69 & 27.18 & 40.39\\

& T-ALIGN 
& 53.94 & 65.49 & \textbf{88.26} & 69.23 
& 27.36 & 67.07 & 70.11 & 54.85
& 20.81 & \textbf{63.57} & 50.24 & 44.87 & 56.32\\
\cdashline{2-15} 

& \textsc{CAGE} 
& 59.39 & \textbf{69.52} & 85.78 & \textbf{71.56}
& \textbf{30.89} & \textbf{71.60} & \textbf{76.48} & \textbf{59.66}
& \textbf{27.45} & 61.38 & \textbf{51.77} & \textbf{46.86} & \textbf{59.36}\\

& $\Delta$
& \down{2.86}& \up{4.03}& \down{2.48}& \up{2.33}
& \up{3.53}& \up{4.53}& \up{6.37}& \up{4.81}
& \up{6.64}& \down{2.19}& \up{1.53}& \up{1.99} &\up{3.04}
\\

\hline
\end{tabular}
\caption{Main results on ASQA, ExpertQA, and ELI5 under the TRUST evaluation framework. Qwen3.5-9B is used as the CMI models, while LLaMA families models of different scales are used as SCR models. $\Delta$ denotes the improvement of CAGE over the strongest baseline.}
\label{tab:main_results_llama_appendix}
\end{table*}

\section{More Experimental results}

\subsection{Experimental results on LLaMA Families}
\label{exp:llama}

Table~\ref{tab:main_results_llama_appendix} demonstrates that CAGE exhibits strong generalization across the LLaMA family. Despite substantial differences in the capacity of the SCR models, CAGE consistently achieves the highest overall Trust performance on ASQA, ExpertQA, and ELI5, indicating that its effectiveness is not restricted to a particular model scale. Specifically, CAGE obtains average Trust scores of 58.60\%, 54.24\%, 60.90\%, and 59.36\% with the 2--7B, 3.2--1B, 3.2--3B, and 3--8B models, respectively, outperforming the strongest baselines by 4.04\%, 4.83\%, 5.63\%, and 3.04\%. Moreover, CAGE achieves the best Trust score across all 12 model--dataset combinations. Its Trust scores remain consistently high on ASQA, ranging from 69.16 to 72.68, while reaching 50.09\%--62.02\% on the more challenging ExpertQA dataset and 43.47\%--48.00\% on ELI5.

The consistent improvements across both compact and relatively large LLaMA models suggest that the effectiveness of CAGE mainly arises from its structured attribution mechanism rather than increased model capacity. Particularly substantial gains are observed on difficult settings, such as ASQA with the 3.2--1B model and ExpertQA with the 3.2--3B model, where CAGE improves Trust by 10.29\% and 13.02\%, respectively. Although CAGE does not achieve the best score on every individual metric, it consistently provides the strongest aggregate Trust performance. This result indicates that the cognitive attribution map helps balance answer correctness, response grounding, and citation quality by organizing claim-specific evidence before generation, thereby improving the robustness and cross-scale generalization of citation-grounded generation on LLaMA models.

\subsection{Evidence-Boundary Overrun Evaluation}
\label{app:ebo_metrics}

We evaluate evidence-boundary overrun (EBO) using the same 948 ASQA questions and the same top-5 retrieved documents for all methods. The evaluated outputs include Full CAGE, previously generated direct-answerer outputs, and newly generated outputs from the GPT-5.5 and Claude Sonnet 4.6 APIs. We use GPT-5.5 to decompose each long-form answer into atomic facts and apply the same TRUE model used for citation-groundedness evaluation to determine whether each fact is supported by the top-5 documents considered jointly. Refusal and empty responses are excluded from the EBO calculation.

Let \(N\) denote the number of valid responses, \(L_i\) the number of atomic facts in response \(i\), \(p_{i,\ell}\) its \(\ell\)-th atomic fact, and \(D_{i,5}\) the corresponding top-5 retrieved documents. We define
\(\operatorname{Entail}(p_{i,\ell},D_{i,5})\in\{0,1\}\), where a value of \(1\) indicates that the atomic fact is supported by the document set.

\paragraph{Atomic-EBO.}
Atomic-EBO measures the proportion of atomic facts across all valid responses that are unsupported by the top-5 documents:
\begin{equation}
\mathrm{Atomic\text{-}EBO}
=
1-
\frac{
\sum_{i=1}^{N}\sum_{\ell=1}^{L_i}
\operatorname{Entail}(p_{i,\ell},D_{i,5})
}{
\sum_{i=1}^{N}L_i
}.
\end{equation}
This metric is a fact-level micro-average. Responses containing more atomic facts therefore receive greater weight.

\paragraph{EBO-Severity.}
EBO-Severity first computes the proportion of unsupported atomic facts within each valid response and then averages it across responses:
\begin{equation}
\mathrm{EBO\text{-}Severity}
=
\frac{1}{N}
\sum_{i=1}^{N}
\left(
1-
\frac{1}{L_i}
\sum_{\ell=1}^{L_i}
\operatorname{Entail}(p_{i,\ell},D_{i,5})
\right).
\end{equation}
Each response receives equal weight, so this metric captures the average extent to which an answer exceeds its evidence boundary.

\paragraph{Response-EBO.}
Response-EBO measures the proportion of valid responses containing at least one unsupported atomic fact:
\begin{equation}
\begin{aligned}
\mathrm{Response\text{-}EBO}
&=
\frac{1}{N}\sum_{i=1}^{N}\mathbb{I}\left[U_i>0\right], \\
U_i
&=
\sum_{\ell=1}^{L_i}
\left(1-\operatorname{Entail}(p_{i,\ell},D_{i,5})\right),
\end{aligned}
\end{equation}
This metric captures the occurrence of evidence-boundary overrun at the response level, regardless of the number of unsupported facts within the response. All three metrics are reported as percentages, and lower values indicate better evidence alignment.

\subsection{Complete results of the Influence of CMI model across different SCR Backbones}
\label{exp: cog}

Table~\ref{tab:SCR} demonstrates that CAGE can effectively enhance diverse downstream LLMs, including both open-source and proprietary models, without requiring modifications to the underlying generators. Across ASQA, ELI5, and ExpertQA, CAGE consistently outperforms end-to-end baselines, achieving the best TRUST scores of 76.62\% with Claude-4.6 on ASQA and 66.43\% with Claude-4.6 on ExpertQA. Notably, when coupled with GPT-5.5 and Claude-4.6, CAGE improves TRUST by 8.47 and 11.10 points on ASQA, and by 14.26 and 11.10 points on ExpertQA, respectively. These gains are primarily attributed to stronger citation grounding, where CAGE achieves F1$_{\mathrm{CG}}$ scores of 94.48\% on ASQA and 90.86\% on ExpertQA with Claude-4.6. Unlike methods that rely on generator-specific optimization, CAGE separates evidence attribution from final response generation, allowing existing LLMs to leverage explicit support structures through a lightweight integration process. These results demonstrate that the induced attribution map provides a model-agnostic evidence alignment interface, enabling heterogeneous LLM backbones to produce more faithful and better-attributed responses while preserving their original generation capabilities. This property is particularly valuable for real-world applications built on proprietary LLM APIs, where model parameters and training pipelines are inaccessible.

\subsection{Experimental results with same LLMs on CMI and SCR stage}
\label{exp: small}
Tables~\ref{tab:main_results_qwen_same} and \ref{tab:main_results_llama_same} demonstrate that CAGE remains effective even when the Cognitive Map Induction Model and Structured Citation Reasoning Model share the same backbone, indicating that the performance gains originate from explicit attribution modeling rather than backbone heterogeneity. Across different scales of Qwen and LLaMA models, CAGE and CAGE$_{\mathrm{DPO}}$ consistently improve TRUST over existing baselines, showing that the induced attribution graph provides additional reasoning capability beyond the original language models.

For Qwen models, CAGE achieves the highest TRUST scores of 65.05\%, 56.38\%, and 41.75\% on ASQA, ExpertQA, and ELI5 with the 1.5B, 3B, and 7B settings, respectively, while CAGE$_{\mathrm{DPO}}$ further improves TRUST to 62.19\%, 68.58\%, and 47.58\% across different scales. Similar trends are observed on LLaMA models, where CAGE obtains consistent gains, achieving TRUST scores of 63.35\%, 66.49\%, and 72.41\% on ASQA with different model scales, and reaching 60.67\% on ExpertQA with LLaMA-8B. Moreover, improvements are particularly evident in citation grounding, with CAGE variants achieving substantial gains in F1$_{\mathrm{GC}}$, demonstrating stronger evidence alignment.

These results indicate that CAGE does not rely on additional model capacity or complementary backbones. Instead, the intermediate attribution graph explicitly organizes answer units and supporting evidence before generation, reducing attribution ambiguity and constraining citation selection. Therefore, even a single backbone model can benefit from the structured attribution process, validating the effectiveness and generality of CAGE as a model-independent framework for citation-grounded generation.

\subsection{Experimental results when proprietary LLM as CMI model}

Table~\ref{tab:cmi_ablation} investigates the impact of different CMI models on CAGE performance while keeping the SCR models unchanged. The results show that stronger CMI models consistently improve the quality of induced attribution maps, leading to better citation-grounded generation across different SCR backbones. In particular, replacing GPT-5-mini with GPT-5.5 as the CMI model improves the average TRUST score across all datasets and SCR models, demonstrating that the quality of intermediate attribution structures directly affects downstream generation.

Quantitatively, GPT-5.5-based CAGE achieves substantial improvements over GPT-5-mini-based CAGE, especially on ASQA, where TRUST increases from 60.23--63.68\% to 67.75--70.88\%. Similar improvements are observed on ELI5 and ExpertQA, with notable gains in F1$_{\mathrm{GC}}$, indicating that stronger CMI models mainly enhance evidence attribution and citation grounding rather than only answer generation. For example, with Qwen2.5-7B as the SCR model, GPT-5.5 improves F1$_{\mathrm{GC}}$ from 72.48\% to 82.66\% on ASQA and from 47.69\% to 64.08\% on ExpertQA.

These results indicate that the CMI stage serves as a critical attribution bottleneck in CAGE. A stronger map induction model provides more accurate answer--evidence alignment, which enables the SCR model to generate better-supported claims regardless of its backbone architecture. Therefore, CAGE benefits not only from stronger generators but also from improving the quality of the intermediate cognitive attribution map.

\section{Metrics.} We use TRUST~\cite{song2025measuring} as the overall metric, which consists of two dimensions: truthfulness and attribution groundedness (Att-Grd.). Truthfulness includes answer calibration \(\mathrm{EM}^{F1}_{\mathrm{AC}}\) and refusal scoring \(\mathrm{F1}_{\mathrm{RG}}\), while Att-Grd. is computed using the TRUE\footnote{https://huggingface.co/google/t5\_xxl\_true\_nli\_mixture} model. Following Song \textit{et al.}, the test set is divided into answerable questions \(A_g\) and unanswerable questions \(A_{ug}\). For \(\mathrm{EM}^{F1}_{\mathrm{AC}}\), it is formalized as:
\begin{equation}
    \mathrm{EM}_{\mathrm{AC}}^{\mathrm{F} 1}=\frac{2 \mathrm{EM}_{\mathrm{AC}}^{\alpha} \cdot \mathrm{EM}_{\mathrm{AC}}^{\beta}}{\mathrm{EM}_{\mathrm{AC}}^{\alpha}+\mathrm{EM}_{\mathrm{AC}}^{\beta}},
\end{equation}
where
\begin{equation}
\begin{aligned}
\mathrm{EM}_{\mathrm{AC}}^{\alpha}
&=
\frac{1}{|A_r|}
\sum_{q_i \in A_g \cap A_r}
\mathrm{EM}_{\mathrm{AC}}^{q_i}, \\
\mathrm{EM}_{\mathrm{AC}}^{\beta}
&=
\frac{1}{|A_g|}
\sum_{q_i \in A_g \cap A_r}
\mathrm{EM}_{\mathrm{AC}}^{q_i},
\end{aligned}
\end{equation}
and \(A_r\) denotes the set of answered questions. The refusal score \(\mathrm{F1}_{\mathrm{RG}}\) is calculated as \(\mathrm{F1}_{\mathrm{RG}}
=
\frac{1}{2}
\left(
\mathrm{F1}_{\mathrm{ref}}
+
\mathrm{F1}_{\mathrm{ans}}
\right),\)
where \(\mathrm{F1}_{\mathrm{ref}}=\frac{2\mathrm{P_{ref}}\cdot \mathrm{R_{ref}}}{\mathrm{P_{ref}}+\mathrm{R_{ref}}}\), \(\mathrm{F1}_{\mathrm{ans}}=\frac{2\mathrm{P_{ans}}\cdot \mathrm{R_{ans}}}{\mathrm{P_{ans}}+\mathrm{R_{ans}}}\), and
\begin{equation}
\begin{aligned}
P_{\mathrm{ref}}
=
\frac{
|\neg A_r \cap \neg A_g|
}{
|\neg A_r|
},
R_{\mathrm{ref}}
=
\frac{
|\neg A_r \cap \neg A_g|
}{
|\neg A_g|
}, \\
P_{\mathrm{ans}}
=
\frac{
| A_r \cap  A_g|
}{
| A_r|
}, 
R_{\mathrm{ans}}
=
\frac{
| A_r \cap  A_g|
}{
| A_g|
}.
\end{aligned}
\end{equation}
For Att-Grd., following~\cite{gao-etal-2023-enabling}, we report the citation-grounded F1 score \(\mathrm{F1_{CG}}=\frac{2\cdot \mathrm{CP} \cdot \mathrm{CR}}{\mathrm{CP} + \mathrm{CR}},\):
where \(\mathrm{CP}\) and \(\mathrm{CR}\) denote citation precision and recall, respectively. Finally, \(\mathbf{TRUST}=\frac{1}{3}(\mathrm{F1_{RG}+EM^{F1}_{AC}+F1_{CG}})\).

\begin{table}[t]
\centering
\footnotesize
\setlength{\tabcolsep}{2.5pt}{

\begin{tabular}{lcccccc}
\toprule
Method & G & A &
$\mathrm{EM}^\text{F1}_{\mathrm{AC}}$ &
$\mathrm{F1}_{\mathrm{RG}}$ &
$\mathrm{F1}_{\mathrm{CG}}$ &
\textbf{TRUST} \\
\midrule

\multicolumn{7}{c}{ASQA} \\
\midrule
\rowcolor{proprietarygray}
Ground-GRPO & \multicolumn{2}{c}{End-to-end} & 47.22 & 62.88 & 81.94 & 64.01 \\
\rowcolor{proprietarygray}
GPT-5.5 & \multicolumn{2}{c}{End-to-end} & 66.18 & 61.54 & 70.69 & 66.14 \\
\rowcolor{proprietarygray}
Claude-4.6 & \multicolumn{2}{c}{End-to-end} & 70.46 & 64.56 & 73.22 & 69.41 \\
CAGE & 9B & 4B & 59.50 & 66.54 & 86.82 & 70.95 \\
CAGE & 9B & 4B-Inst. & 61.07 & 68.34 & 88.10 & 72.50 \\
CAGE & 9B & 4B-Think. & 61.03 & 67.07 & 86.07 & 71.39 \\
CAGE & 9B & GPT-5.5 & 61.41 &  70.09 & 92.32 & \textbf{74.61} \\
CAGE & 9B & Claude-4.6 & 65.29 &  70.09 & 94.48 & \textbf{76.62} \\
\midrule

\multicolumn{7}{c}{ELI5} \\
\midrule
\rowcolor{proprietarygray}
Ground-GRPO & \multicolumn{2}{c}{End-to-end} & 20.32 & 61.85 & 51.71 & 44.62 \\
\rowcolor{proprietarygray}
GPT-5.5 & \multicolumn{2}{c}{End-to-end} & 42.76 & 45.81 & 40.44 & 43.00 \\
\rowcolor{proprietarygray}
Claude-4.6 & \multicolumn{2}{c}{End-to-end} & 37.33 & 61.03 & 46.91 & 48.42 \\
CAGE & 9B & 4B & 26.98 & 61.11 & 57.86 & 48.65 \\
CAGE & 9B & 4B-Inst. & 30.73 & 61.48 & 57.58 & \textbf{49.92} \\
CAGE & 9B & 4B-Think. & 27.48 & 59.03 & 58.26 & 48.26 \\

\midrule
\multicolumn{7}{c}{ExpertQA} \\
\midrule
\rowcolor{proprietarygray}
Ground-GRPO & \multicolumn{2}{c}{End-to-end} & 20.56 & 67.14 & 59.37 & 49.02 \\
\rowcolor{proprietarygray}
GPT-5.5 & \multicolumn{2}{c}{End-to-end} & 46.74 & 54.46 & 49.78 & 50.33 \\
\rowcolor{proprietarygray}
Claude-4.6 & \multicolumn{2}{c}{End-to-end} & 47.66 & 64.54 & 53.80 & 55.33 \\
CAGE & 9B & 4B & 36.27 & 62.22 & 78.81 & 62.41 \\
CAGE & 9B & 4B-Inst. & 36.61 & 73.35 & 80.41 & 63.46 \\
CAGE & 9B & 4B-Think. & 37.59 & 73.24 & 79.15 & 63.33 \\
CAGE & 9B & GPT-5.5 & 32.88 &  71.74 & 89.15 & \textbf{64.59} \\ 
CAGE & 9B & Claude-4.6 & 36.70 &  71.74 & 90.86 & \textbf{66.43} \\
\bottomrule
\end{tabular}}
\caption{Backbone capability and proprietary model comparison. G and A denote the map induction model and SCR model, respectively. 9B denotes Qwen3.5-9B; Q4B, 4B-Inst., and 4B-Think. denote Qwen3-4B, Qwen3-4B-Instruct-2507, and Qwen3-4B-Thinking-2507, respectively. CAGE fixes Qwen3.5-9B as G and evaluates Qwen3-4B variants as A, while gray rows denote end-to-end baselines.}
\label{tab:SCR}
\end{table}

\begin{table*}[h]
\centering
\small
\setlength{\tabcolsep}{6pt} 
\renewcommand{\arraystretch}{1.2}
\begin{tabular}{
cc
cccc
cccc
cccc
}
\toprule
\multirow{3}{*}{Size} & \multirow{3}{*}{Type} 
& \multicolumn{4}{c}{\makecell{ASQA(610 answerable,\\338 unanswerable)}}
& \multicolumn{4}{c}{\makecell{EXPERTQA(682 answerable,\\1487 unanswerable)}} 
& \multicolumn{4}{c}{\makecell{ELI5(207 answerable,\\793 unanswerable)}} \\

\cline{3-14}

& & \multicolumn{2}{c}{Truthfulness} & \multicolumn{1}{c}{AG} & \multirow{2}{*}{Trust} 
  & \multicolumn{2}{c}{Truthfulness} & \multicolumn{1}{c}{AG} & \multirow{2}{*}{Trust} 
  & \multicolumn{2}{c}{Truthfulness} & \multicolumn{1}{c}{AG} & \multirow{2}{*}{Trust} \\
\cline{3-5} \cline{7-9} \cline{11-13}

& & F1$_\text{AC}$ & F1$_\text{GR}$ & F1$_\text{GC}$ & 
  & F1$_\text{AC}$ & F1$_\text{GR}$ & F1$_\text{GC}$ & 
  & F1$_\text{AC}$ & F1$_\text{GR}$ & F1$_\text{GC}$ & \\
\midrule

\multirow{7}{*}{1.5b}
& ICL 
& 50.55 & 41.74 & 6.69 & 32.99 
& 30.67 & 26.09 & 6.89 & 21.22
& 20.56 & 17.78 & 4.99 & 14.44 \\

& PostCite 
& 16.36 & 52.46 & 15.40 & 28.07 
& 22.22 & 48.66 & 16.92 & 29.27
& 15.63 & 26.71 & 5.17 & 15.84 \\

& PostAttr 
& 16.36 & 52.46 & 4.45 & 24.42 
& 22.22 & 48.66 & 13.15 & 28.01
& 15.63 & 26.71 & 0.62 & 14.32 \\

& FRONT 
& \textbf{57.74} & 41.36 & 55.70 & 51.60 
& \textbf{29.15} & 24.60 & 50.22 & 34.66
& 19.57 & 17.29 & 37.70 & 24.85 \\

& T-ALIGN 
& 52.68 & 62.38 & 66.81 & 60.62
& 25.06	& \textbf{68.38}	& 51.44	& 48.29
& 19.03 & \textbf{57.91} & 31.63 & 36.19 \\
\cdashline{2-14} 
& \textsc{CAGE} 
& 50.71 & \textbf{66.89} & \textbf{77.55} & \textbf{65.05}
& 24.80	& 52.79	& 53.97	& 43.85
& 21.98 & 43.31 & 33.89 & 33.07 \\

& CAGE$_{\text{DPO}}$ 
& 47.81 & 64.25 & 74.49 & 62.19
& 24.53	& 63.91	& \textbf{61.48} & \textbf{49.98}
& \textbf{24.69} & 56.32 & \textbf{38.24} & \textbf{39.78} \\

\midrule

\multirow{7}{*}{3b}
& ICL 
& 37.72 & 51.36 & 51.72 & 46.93 
& 35.14 & 49.65 & 42.67 & 42.49
& 29.12 & 46.31 & 34.34 & 36.59 \\

& PostCite 
& 9.58 & 35.30 & 10.94 & 18.61 
& 0 & 40.66 & 0 & 13.55
& 21.73 & 48.49 & 7.56 & 25.93 \\

& PostAttr 
& 9.58 & 35.30 & 36.29 & 27.06 
& 0 & 40.66 & 0 & 13.55
& 21.73 & 48.49 & 1.31 & 23.84 \\

& FRONT 
& 55.15 & 44.01 & 62.72 & 53.96 
& 25.67 & 29.86 & 44.48 & 33.34
& 18.69 & 25.37 & 37.40 & 27.15 \\

& T-ALIGN 
& 55.19 & 63.76 & 78.64 & 65.86
& 20.97	& 65.79	& 60.25	& 49.0
& 22.52 & 64.38 & 42.01 & 42.97 \\
\cdashline{2-14} 
& \textsc{CAGE} 
& \textbf{55.78} & 61.12 & 78.09 & 65.00
& \textbf{31.71} & \textbf{68.17} & 69.28 & \textbf{56.38}
& 25.12 & 58.80 & 41.32 & 41.75 \\

& CAGE$_{\text{DPO}}$ 
& 53.99 & \textbf{66.81} & \textbf{84.96} & \textbf{68.58}
& 30.24	& 67.63	& \textbf{70.27} & 56.05
& \textbf{28.41} & \textbf{61.16} & \textbf{48.34} & \textbf{45.97} \\

\midrule

\multirow{7}{*}{7b}
& ICL 
& 58.94 & 54.34 & 75.46 & 62.91 
& 36.33 & 42.28 & 56.09 & 44.9
& 28.27 & 37.13 & 44.13 & 36.51 \\

& PostCite 
& 27.52 & 45.93 & 4.19 & 25.88 
& 25.58 & 54.9 & 13.77 & 31.42
& 21.82 & 22.23 & 7.03 & 17.03 \\

& PostAttr 
& 27.52 & 45.93 & 17.92 & 30.46 
& 25.58 & 54.9 & 12.46 & 30.98
& 21.82 & 22.23 & 0.96 & 15.00 \\

& FRONT 
& 64.58 & 60.08 & 58.27 & 60.98 
& 32.41 & 55.56 & 67.35 & 51.77
& 28.27 & 54.14 & 56.61 & 46.34 \\

& T-ALIGN 
& 55.04 & \textbf{66.22} & 83.57 & 68.28
& 25.57	& 69.16	& 62.7	& 52.48
& 24.30 & 63.79 & 47.02 & 45.04 \\
\cdashline{2-14} 
& \textsc{CAGE} 
& 60.24 & 59.03 & 82.15 & 67.14
& 34.78 & 69.00 & 71.19 & 58.32
& 28.84 & 54.48 & 48.80 & 44.04 \\

& CAGE$_{\text{DPO}}$ 
& \textbf{61.04} & 65.93 & \textbf{84.44} & \textbf{70.47}
& \textbf{35.20} & \textbf{72.95} & \textbf{78.57}	& \textbf{62.23}
& \textbf{31.59} & \textbf{61.86} & \textbf{49.45} & \textbf{47.58} \\

\bottomrule
\end{tabular}
\caption{Experimental results of different model scales of Qwen on the ASQA, QAMPARI, and ELI5 datasets.}
\label{tab:main_results_qwen_same}
\end{table*}

\begin{table*}[t]
\centering
\small
\setlength{\tabcolsep}{6pt} 
\begin{tabular}{
cc
cccc
cccc
cccc
}
\toprule
\multirow{3}{*}{Size} & \multirow{3}{*}{Type} 
& \multicolumn{4}{c}{\makecell{ASQA(610 answerable,\\338 unanswerable)}}
& \multicolumn{4}{c}{\makecell{EXPERTQA(682 answerable,\\1487 unanswerable)}} 
& \multicolumn{4}{c}{\makecell{ELI5(207 answerable,\\793 unanswerable)}} \\

\cline{3-14}

& & \multicolumn{2}{c}{Truthfulness} & \multicolumn{1}{c}{AG} & \multirow{2}{*}{Trust} 
  & \multicolumn{2}{c}{Truthfulness} & \multicolumn{1}{c}{AG} & \multirow{2}{*}{Trust} 
  & \multicolumn{2}{c}{Truthfulness} & \multicolumn{1}{c}{AG} & \multirow{2}{*}{Trust} \\
\cline{3-5} \cline{7-9} \cline{11-13}

& & F1$_{AC}$ & F1$_{GR}$ & F1$_{GC}$ & 
  & F1$_{AC}$ & F1$_{GR}$ & F1$_{GC}$ & 
  & F1$_{AC}$ & F1$_{GR}$ & F1$_{GC}$ & \\
\midrule

\multirow{7}{*}{2-7b}
& ICL 
& 0.00 & 26.28 & 0.00 & 8.76
& 0.00 & 41.01 & 9.25 & 16.84
& 0.50 & 46.71 & 5.04 & 15.57 \\

& PostCite 
& 0.07 & 35.23 & 0.00 & 11.77
& 4.85 & 44.27 & 5.23 & 18.12
& 1.86 & 44.98 & 13.80 & 17.29 \\

& PostAttr 
& 0.07 & 35.23 & 0.00 & 11.77
& 4.85 & 44.27 & 2.26 & 17.13
& 1.86 & 44.98 & 0.00 & 15.61 \\

& FRONT 
& 60.47 & 39.15 & 68.86 & 56.16
& 9.33 & 23.92 & 74.75 & 36.00
& 21.66 & 17.15 & 52.72 & 30.51 \\

& T-ALIGN 
& 52.48 & 66.12 & 83.94 & 67.51 
& 25.03 & 67.91 & 62.46 & 51.8
& 22.54 & 63.27 & 47.35 & 44.39 \\

& \textsc{CAGE} 
& 49.71 & 63.12 & 77.25 & 63.35
& 23.88 & 64.09 & 78.45 & 55.48
& 22.49 & 57.89 & 40.19 & 40.19 \\

& CAGE$_{\text{DPO}}$ 
& 50.34 & 62.72 & 83.86 & 65.64
& 22.15 & 62.55 & 74.35 & 53.01
& 19.47 & 57.09 & 40.47 & 39.01 \\

\hline

\multirow{7}{*}{3.2-1b}
& ICL 
& 35.95 & 50.94 & 9.96 & 32.28
& 21.55 & 32.83 & 9.04 & 21.14
& 12.87 & 27.10 & 5.23 & 15.07 \\

& PostCite 
& 0.59 & 50.22 & 0.24 & 17.02
& 5.48 & 49.1 & 2.67 & 19.08
& 2.04 & 50.88 & 1.02 & 17.98 \\

& PostAttr 
& 0.48 & 48.42 & 0.00 & 16.30
& 8.24 & 47.72 & 1.5 & 19.15
& 2.04 & 50.88 & 0.07 & 17.66 \\

& FRONT 
& 48.22 & 54.48 & 48.29 & 50.33
& 20.83 & 29.26 & 37.45 & 29.18
& 16.11 & 20.76 & 30.19 & 22.35 \\

& T-ALIGN 
& 38.64 & 58.61 & 7935 & 58.87
& 20.32 & 64.87 & 62.1 & 49.1
& 13.20 & 59.35 & 48.21 & 40.25 \\

& \textsc{CAGE} 
& 43.97 & 58.77 & 67.51 & 56.75
& 24.59	& 64.84	& 62.41	& 50.62
& 13.13 & 56.54 & 36.56 & 35.41 \\

& CAGE$_{\text{DPO}}$ 
& 58.59 & 69.62 & 83.34 & 70.52
& 25.67 & 65.21 & 62.95 & 51.28
& 13.71 & 54.29 & 35.17 & 34.39 \\

\hline

\multirow{7}{*}{3.2-3b}
& ICL 
& 2.04 & 27.98 & 53.95 & 27.99
& 33.5 & 51.21 & 38.37 & 41.03
& 18.55 & 55.56 & 30.70 & 34.94 \\

& PostCite 
& 31.03 & 56.59 & 2.29 & 36.87
& 25.68 & 38.11 & 5.29 & 23.03
& 18.12 & 25.14 & 4.44 & 15.90 \\

& PostAttr 
& 29.76 & 56.71 & 4.69 & 30.39
& 25.45 & 38.58 & 3.4 & 22.48
& 18.48 & 25.14 & 0.53 & 14.72 \\

& FRONT 
& 63.19 & 49.45 & 57.46 & 56.70
& 27.24 & 43.34 & 50.91 & 40.5
& 19.95 & 32.21 & 41.97 & 31.38 \\

& T-ALIGN 
& 59.82 & 66.38 & 84.21 & 70.14
& 11.72 & 56.93 & 78.35 & 49.0
& 18.33 & 62.79 & 55.87 & 45.66 \\

& \textsc{CAGE} 
& 52.31 & 60.47 & 86.70 & 66.49
& 24.86 & 63.99 & 82.19 & 57.01
& 29.43 & 61.85 & 45.64 & 45.64 \\

& CAGE$_{\text{DPO}}$ 
& 48.14 & 59.26 & 86.75 & 64.72
& 21.25 & 59.62 & 82.58 & 54.49
& 28.51 & 63.84 & 47.47 & 46.61 \\

\hline

\multirow{7}{*}{2-8b}
& ICL 
& 3.01 & 28.58 & 86.50 & 39.36
& 2.82 & 42.5 & 69.46 & 38.26
& 0.00 & 44.23 & 0.00 & 14.74 \\

& PostCite 
& 32.98 & 53.31 & 28.01 & 38.10
& 14.06 & 50.08 & 7.09 & 23.74
& 20.80 & 45.88 & 8.06 & 24.91 \\

& PostAttr 
& 32.98 & 53.31 & 5.95 & 30.75
& 14.06 & 50.08 & 6.29 & 23.47
& 20.80 & 45.88 & 1.25 & 22.64 \\

& FRONT 
& 62.25 & 41.62 & 66.14 & 56.67
& 30.34 & 24.92 & 56.7 & 37.32
& 18.99 & 17.85 & 44.69 & 27.18 \\

& T-ALIGN 
& 53.94 & 65.49 & 88.26 & 69.23 
& 27.36 & 67.07 & 70.11 & 54.85
& 20.81 & 63.57 & 50.24 & 44.87 \\

& \textsc{CAGE} 
& 61.38 & 69.93 & 85.92 & 72.41
& 31.39 & 71.15 & 79.46 & 60.67
& 27.45 & 61.38 & 51.77 & 46.86 \\

& CAGE$_{\text{DPO}}$ 
& 58.59 & 69.62 & 83.34 & 70.52
& 31.25 & 71.24 & 75.19 & 59.22
& 27.61 & 61.44 & 49.15 & 46.07 \\

\bottomrule
\end{tabular}
\caption{Experimental results of different model scales of Llama on the ASQA, QAMPARI, and ELI5 datasets.}
\label{tab:main_results_llama_same}
\end{table*}

\section{Case Study}

Figure~\ref{fig:into_thin_air_case} presents a case study illustrating how CAGE performs evidence-grounded reasoning through attribution graph construction. Given a complex question requiring multi-source information integration, CAGE first identifies multiple support subgraphs from retrieved documents, where each subgraph corresponds to a distinct answer unit with explicit evidence paths. In this example, the induced graph separates different aspects of the answer, including Anatoli Boukreev's death, Scott Fischer's death, and the total number of casualties, while linking each answer unit to its supporting documents. This decomposition prevents unsupported information mixing and enables precise claim-level citation generation. Compared with directly generating answers from retrieved documents, CAGE explicitly determines which evidence supports each answer unit before decoding, resulting in more faithful responses with well-aligned citations. Notably, CAGE does not simply select all death-related documents; instead, it distinguishes different evidence-supported answer units and attributes each claim to its corresponding source.

\begin{table*}[t]
\centering
\small
\setlength{\tabcolsep}{10pt}

\begin{tabular}{cc|cccc|cccc}
\toprule
\multirow{2}{*}{Dataset}
& \multirow{2}{*}{SCR Model}
& \multicolumn{4}{c|}{GPT-5-mini as CMI}
& \multicolumn{4}{c}{GPT-5.5 as CMI} \\
\cline{3-10}

& 
& $\mathrm{EM}^\text{F1}_\text{AC}$ 
& F1$_\text{GR}$ 
& F1$_\text{GC}$ 
& TRUST
& $\mathrm{EM}^\text{F1}_\text{AC}$ 
& F1$_\text{GR}$ 
& F1$_\text{GC}$ 
& TRUST \\

\midrule

\multirow{4}{*}{ASQA}
& Llama3.2-1B
& 63.15 & 52.13 & 65.40 & 60.23
& 67.79 & 61.71 & 73.76 & 67.75 \\

& Llama3.2-3B
& 65.49 & 52.36 & 73.20 & 63.68
& 68.52 & 61.71 & 81.01 & 70.42 \\

& Qwen2.5-3B
& 64.02 & 52.13 & 71.76 & 62.64
& 67.84 & 61.71 & 78.99 & 69.51 \\

& Qwen2.5-7B
& 65.15 & 52.58 & 72.48 & 63.40
& 68.26 & 61.71 & 82.66 & 70.88 \\

\midrule

\multirow{4}{*}{ELI5}
& Llama3.2-1B
& 23.39 & 43.15 & 22.93 & 29.82
& 24.99 & 40.70 & 42.94 & 36.21 \\

& Llama3.2-3B
& 25.03 & 43.32 & 31.04 & 33.13
& 28.95 & 40.99 & 56.19 & 42.04 \\

& Qwen2.5-3B
& 25.34 & 43.24 & 27.81 & 32.13
& 29.55 & 41.09 & 53.83 & 41.49 \\

& Qwen2.5-7B
& 25.41 & 43.24 & 30.15 & 32.93
& 30.26 & 41.09 & 55.08 & 42.14 \\

\midrule

\multirow{4}{*}{ExpertQA}
& Llama3.2-1B
& 33.66 & 53.64 & 37.69 & 41.67
& 32.17 & 45.01 & 53.18 & 43.45 \\

& Llama3.2-3B
& 37.88 & 55.18 & 48.49 & 47.18
& 36.82 & 45.03 & 64.75 & 48.87 \\

& Qwen2.5-3B
& 38.53 & 55.08 & 46.56 & 46.72
& 35.38 & 45.08 & 62.94 & 47.80 \\

& Qwen2.5-7B
& 39.19 & 55.18 & 47.69 & 47.35
& 37.54 & 45.08 & 64.08 & 48.90 \\

\bottomrule
\end{tabular}

\caption{Effect of CMI model quality on CAGE performance with different SCR backbones. GPT-5-mini and GPT-5.5 are used as CMI models, while finetuned Qwen and LLaMA models serve as SCR models.}
\label{tab:cmi_ablation}
\end{table*}

\begin{figure*}[t]
\centering

\begin{tcolorbox}[
title=\textbf{Case Study},
colback=gray!3,
colframe=black,
boxrule=0.6pt,
width=\textwidth
]

\textbf{Question}

Who died in the book \emph{Into Thin Air}?

\vspace{0.5em}
\hrule
\vspace{0.8em}

\begin{minipage}[t]{0.52\textwidth}

\textbf{Retrieved Documents}

\vspace{0.5em}

\textbf{D1}

Many public criticisms at Krakauer concerning the accuracy of each man's account of what happened on the mountain during the 1996 climbs.  \textbf{Boukreev was killed in 1997 in an avalanche during a winter ascent of Annapurna in Nepal.}

\vspace{0.5em}

\textbf{D2}

Journalist Jon Krakauer, on assignment from \textit{Outside} magazine, was in a party led by guide Rob Hall that lost four climbers on the south side; he afterwards published the bestseller \"Into Thin Air\" (1997), which related his experience.  \textbf{Anatoli Boukreev was a guide in Scott Fischer's party, which lost Scott Fischer, but no clients.}

\vspace{0.5em}

\textbf{D3}

\emph{Into Thin Air} details the author's experience at the 1996 Mount Everest disaster,  \textbf{in which eight climbers were killed and several others were stranded by a rogue storm.}

\vspace{0.5em}

\textbf{D4}

Adventure Consultants was the guiding company led by Rob Hall.Members of his party died on the way down. The disaster became very well known, with ten million people reading about it in \emph{Into Thin Air}. In 1996, Hall also employed two Sherpa people who managed to survive.

\vspace{0.5em}

\textbf{D5}

The storm led to a number of deaths,  \textbf{including both head guides (Rob Hall and Scott Fischer)}. Krakauer clarified his initial statements—especially those regarding the death of Andy Harris—in \emph{Into Thin Air}.

\end{minipage}
\hfill
\begin{minipage}[t]{0.44\textwidth}

\textbf{Support Graph}

\vspace{1em}

\centering

\begin{tikzpicture}[
    node distance=0.9cm,
    >=Latex,
    every node/.style={font=\small}
]

\begin{scope}
\node[draw, rounded corners, fill=blue!15] (Q0) {Q};
\node[draw, fill=green!15, below=of Q0] (D1) {D1};
\node[draw, fill=orange!20, below=of D1] (A1) {A$_1$};
\draw[->, thick] (Q0) -- (D1);
\draw[->, thick] (D1) -- (A1);
\end{scope}

\begin{scope}[xshift=2.5cm]
\node[draw, rounded corners, fill=blue!15] (Q1) {Q};
\node[draw, fill=green!15, below left=0.9cm and 0.5cm of Q1] (D2) {D2};
\node[draw, fill=green!15, below right=0.9cm and 0.5cm of Q1] (D5) {D5};
\node[draw, fill=orange!20, below=2.3cm of Q1] (A2) {A$_2$};
\draw[->, thick] (Q1) -- (D2);
\draw[->, thick] (Q1) -- (D5);
\draw[->, thick] (D2) -- (A2);
\draw[->, thick] (D5) -- (A2);
\end{scope}

\begin{scope}[xshift=5cm]
\node[draw, rounded corners, fill=blue!15] (Q2) {Q};
\node[draw, fill=green!15, below=of Q2] (D3) {D3};
\node[draw, fill=orange!20, below=of D3] (A3) {A$_3$};
\draw[->, thick] (Q2) -- (D3);
\draw[->, thick] (D3) -- (A3);
\end{scope}

\end{tikzpicture}

\vspace{0.6em}

\raggedright

\textbf{Edges:}

\emph{subgraph0}

nodes:Q  inputs: N/A  outputs: D1

nodes:D1  inputs: Q  outputs: A1

nodes:A1  inputs: D1  outputs: N/A

subgraph\_description: The document supports the candidate answer by confirming that Anatoli Boukreev died in 1997 during a winter ascent of Annapurna.

\emph{subgraph1}

nodes:Q  inputs: N/A  outputs: D2,D5

nodes:D2  inputs: Q  outputs: A2

nodes:D5  inputs: Q  outputs: A2

nodes:A2  inputs: D2,D5  outputs: N/A

subgraph\_description: The documents support that Scott Fischer died in the 1996 Everest disaster, as they mention his party and the tragic event.

\emph{subgraph2}

nodes:Q  inputs: N/A  outputs: D3

nodes:D3  inputs: Q  outputs: A3

nodes:A3  inputs: D3  outputs: N/A

subgraph\_description: The document supports the candidate answer by stating that eight climbers were killed during the 1996 Mount Everest disaster.

\end{minipage}

\vspace{0.8em}

\hrule

\vspace{0.8em}

\textbf{Generated Answer}

Anatoli Boukreev, who served as a guide on the 1996 Mount Everest expedition described in Jon Krakauer's \emph{Into Thin Air}, died in 1997 during a winter ascent of Annapurna in Nepal [1]. Scott Fischer, a guide in the 1996 Mount Everest disaster, died during the 1996 Everest disaster while leading his own expedition [2][5]. Eight climbers died during the 1996 Mount Everest disaster, as documented in Jon Krakauer's book \emph{Into Thin Air} [3].

\end{tcolorbox}

\caption{
A case study illustrating graph-based reasoning for multi-source information integration.
Subgraph 0 identifies Anatoli Boukreev's death,
Subgraph 1 captures Scott Fischer's death supported by two documents,
and Subgraph 2 provides the total death toll of eight climbers.
}
\label{fig:into_thin_air_case}
\end{figure*}


\begin{figure*}[t]
\centering
\begin{GraphPromptBox}
  Important rules for subgraph construction:
  1. Each subgraph corresponds to exactly one answer node: subgraph0 -> A1, subgraph1 -> A2, subgraph2 -> A3, and so on. Never reuse the same A node across multiple subgraphs.
  2. Each subgraph expresses one semantically independent answer aspect. If different documents support different facts or angles of the answer, split them into separate subgraphs with separate A nodes.
  3. Only group multiple T nodes into the same subgraph when they jointly support the same single answer aspect (as T1 and T5 both support who won the 2018 World Cup in subgraph0). 
  The most important: 4. When all documents fail to support answering questions, an empty chart should be output. In the empty chart, the input and output of each document node are empty, and Reason (Within 20 tokens) should provide an explanation for each document indicating that it is not supported.
  Empty edges is:
  Edges:
  nodes:Q       inputs:         outputs:
  nodes:T1      inputs:         outputs:
  reason: ...
  nodes:T2      inputs:         outputs:
  reason: ...
  nodes:T3      inputs:         outputs:
  reason: ...
  nodes:T4      inputs:         outputs:
  reason: ...
  nodes:T5      inputs:         outputs:
  reason: ...
  For example, the input is:
  Who was the champion of the 2018 World Cup?
  Document 1: France defeated Croatia 4-2 in the 2018 final.
  Document 2: Kylian Mbappé contributed 2 goals in the final, playing a significant role in France's championship victory.
  Document 3: It discusses the 2022 World Cup final
  Document 4: It discusses the semi-finals of the 2018 World Cup
  Document 5: France was the champion of the 2018 World Cup
  We want the model output(one subgraph, one A node, one independent answer aspect.):
  Edges
  subgraph0
  nodes:Q       inputs: N/A     outputs: T1,T5
  nodes:T1      inputs: Q       outputs: A1
  nodes:T5      inputs: Q       outputs: A1
  nodes:A1      inputs: T1, T5  outputs: N/A
  subgraph_description: Document 1 and Document 5 jointly directly support the answer, explicitly stating that France was the champion of the 2018 World Cup. Document 1 provides details of the final score, while Document 5 directly states the ownership of the championship

Based on a given single question node Q and several document title nodes T (T1–T5), construct one or more causal subgraphs containing answer node A through causal reasoning.  The question should be treated as answerable if at least one document node contains explicit statements or clear indicators that directly answer the question, even if other documents are irrelevant.  When the question involves multiple entities, multi-hop reasoning, or requires information decomposition, multiple subgraphs can be generated, but they must maintain logical clarity and semantic independence. The A node in the generated edges should be inferred from the supporting T node(s). The description of the A node should indicate that all its input T nodes individually support the A node, or collectively contribute to the generation of the A node. T nodes that are irrelevant to the A node should not be included as inputs to the A node.
  {user_input}
\end{GraphPromptBox}
\end{figure*}


\begin{figure*}[t]
\centering
\begin{AnswerPromptBox}
Generate the final answer from the retrieved documents, using the graph only to decide citations.

  Graph schema:
  - Q is the user question.
  - T nodes are retrieved documents. T1 maps to citation [1], T2 maps to citation [2], and so on.
  - Each subgraph is one citation plan. It lists the T documents that should support one answer sentence.
  - subgraph_description is only a hint about why those T documents were selected; do not copy it as the answer.

  Rules:
  - If at least one evidence path is provided, answer the question. Do not refuse.
  - For each subgraph, write exactly one independent answer sentence grounded only in the documents used by that
  subgraph.
  - Do not merge subgraphs, even if they appear to support the same answer.
  - Keep each sentence short and focused on one factual claim.
  - Cite that sentence with exactly the numeric citations for the T nodes in that subgraph, unless a listed document
  does not support the sentence.
  - Use document text, not subgraph_description text, to decide the factual content of the sentence.
  - Every answer sentence must end with numeric citations in this exact form: sentence text [1].
  - Do not put citations on a separate line or at the start of a sentence.
  - Do not cite documents that are not listed for the evidence paths.
  - Output the independent answer sentences as one concise paragraph.

  Question:
  {question}

  Evidence paths extracted from the graph:
  Path {i}:
  - Citation nodes for one answer sentence: {[citation_ids]}
  - Selection hint, not answer text: {subgraph_description}

  Referenced documents:
  [{citation_id}] {document_title}
  {document_text}

  Original graph structure:
  {graph_input}
\end{AnswerPromptBox}
\end{figure*}

\end{document}